\documentclass[twoside,11pt]{article}

\usepackage[abbrvbib, preprint]{jmlr2e}

\newcommand*{\colonEquals}{\mathrel{\rlap{%
                     \raisebox{0.3ex}{$\cdot$}}%
                     \raisebox{-0.3ex}{$\cdot$}}%
                     =}

\newcommand{\rk}{Runge-Kutta 4}

\newcommand{\anode}{\citet{gholami2019anode}}
\newcommand{\dto}{Disc-Opt}
\newcommand{\otd}{Opt-Disc}

\newcommand{\regL}{L}
\newcommand{\calJ}{\mathcal{J}}
\newcommand{\calG}{\mathcal{G}}
\newcommand{\calL}{\mathcal{L}}
\newcommand{\calR}{\mathcal{R}}

\newcommand{\bfth}{\boldsymbol{\theta}}

\newcommand{\myD}[1]{\mathbf{D}_{\mathbf{\theta}_{#1}}}

\newcommand{\bfu}{\mathbf{u}}

\newcommand{\bfg}{\mathbf{g}}
\newcommand{\bfv}{\mathbf{v}}
\newcommand{\bfy}{\mathbf{y}}
\newcommand{\bfz}{\mathbf{z}}

\newcommand{\bfA}{\mathbf{A}}

\newcommand{\bfF}{\mathbf{F}}

\def\du{\ensuremath{\mathrm{d}}}

\jmlrheading{}{}{}{}{}{}{Derek Onken and Lars Ruthotto}

\ShortHeadings{Disc-Opt vs. Opt-Disc for Time-Series Regression and CNF}{Onken and Ruthotto}
\firstpageno{1}

\usepackage[utf8]{inputenc}
\usepackage{graphicx}
\usepackage{booktabs}
\usepackage{siunitx} 
\usepackage{natbib}
\usepackage{mathtools} 
\usepackage{multirow}
\usepackage{xcolor}
\usepackage{times}
\usepackage{subfig}

\begin{document}

\title{Discretize-Optimize vs. Optimize-Discretize for Time-Series Regression and Continuous Normalizing Flows}

\author{\name Derek Onken \email donken@emory.edu\\
       \name Lars Ruthotto \email lruthotto@emory.edu \\
       \addr Department of Computer Science\\
       Department of Mathematics\\
       Emory University\\
       Atlanta, GA 30322, USA}

\editor{}

\maketitle

\begin{abstract}%
	We compare the discretize-optimize (Disc-Opt) and optimize-discretize (Opt-Disc) approaches for time-series regression and continuous normalizing flows (CNFs) using neural ODEs.
	Neural ODEs are ordinary differential equations (ODEs) with neural network components.
	Training a neural ODE is an optimal control problem where the weights are the controls and the hidden features are the states.
	Every training iteration involves solving an ODE forward and another backward in time, which can require large amounts of computation, time, and memory. 
	Comparing the Opt-Disc and Disc-Opt approaches in image classification tasks, Gholami et al.~(2019) suggest that Disc-Opt is preferable due to the guaranteed accuracy of gradients. 
	In this paper, we extend the comparison to neural ODEs for time-series regression and CNFs. Unlike in classification, meaningful models in these tasks must also satisfy additional requirements beyond accurate final-time output, e.g., the invertibility of the CNF.
	Through our numerical experiments, we demonstrate that with careful numerical treatment, Disc-Opt methods can achieve similar performance as Opt-Disc at inference with drastically reduced training costs. Disc-Opt reduced costs in six out of seven separate problems with training time reduction ranging from 39\% to 97\%, and in one case, Disc-Opt reduced training from nine days to less than one day.
\end{abstract}%
\begin{keywords}%
  neural ODEs, optimal control, Optimize-Discretize, Discretize-Optimize, normalizing flows
\end{keywords}

\section{Introduction}
	
	Coined by \citet{chen2018neural}, neural ODEs are ordinary differential equations (ODEs) with neural network components (Section~\ref{sec:background}).
	Neural ODEs were developed as continuous limits of discrete Residual Networks (ResNets)~\citep{he2016deep}, which can be seen as forward Euler discretizations of a continuous time-dependent ODE~\citep{weinan2017,HaberRuthotto2017}.
	Since ResNets demonstrate impressive performance in applications ranging across image classification, segmentation~\citep{zhang2017segmentation}, and deblurring~\citep{nah2017deblurring}, this result motivated several follow-up works. 
	Examples include continuous models for PolyNet, FractalNet, and RevNet~\citep{lu2017beyond}, the use of higher-order ODE solvers~\citep{Zhu2018}, architectures for convolutional ResNets motivated from partial differential equations (PDEs)~\citep{RuthottoHaber2019}, and continuous versions of normalizing flows \citep{rezende2015,chen2018neural}.
	The last work shows neural ODEs' potential for scientific machine learning applications that involve dynamical systems, which differ from classification and other data science problems traditionally solved with ResNets. This interplay between physics, differential equations, and neural networks is apparent in~\citet{raissi2018deep},~\citet{brunton2019machine},~\citet{kohler2019equivariant}, and~\citet{rackauckas2020universal}.
	We note earlier versions of continuous neural networks reminiscent of neural ODEs can be found in~\citet{RICOMARTINEZ:2007gt} and~\citet{RICOMARTINEZ:1998gt}, which also show the benefits of continuous models for time sequence data and solutions of partial differential equations.

	Training neural ODEs consists of minimizing a regularized loss over the network weights subject to the nonlinear ODE constraint.
	Therefore, training can be seen as an optimal control problem.
	Applying concepts from optimal control theory has emerged as an active research area and more insight has been gained in recent years.
	For example, Pontryagin's maximum principle has been used to efficiently train networks with discrete weights~\citep{li2017maximum}, multigrid methods have been proposed to parallelize forward propagation during training~\citep[e.g.,][]{GuntherEtAl2018}, and analyzing the convergence on the continuous and discrete level has led to novel architectures~\citep{benning2019deep}.

	In the language of optimal control, the adaptive adjoint-based approaches in \citet{chen2018neural}, \citet{rackauckasBlogPost}, and~\citet{grathwohl2018ffjord} can be viewed as Optimize-Discretize (\otd{}) approaches since they optimize the continuous ODE and discretize the optimal dynamics after training.
	A well-known alternative is the Discretize-Optimize (\dto{}) approach, where one first discretizes the continuous problem and then solves a finite-dimensional optimization problem~\citep{gunzburger2003perspectives}. 
	\anode{} give a thorough discussion of the trade-off between Opt-Disc and Disc-Opt as well as numerical examples for neural ODEs in image classification.
	Our goal in this paper is to extend this discussion and perform similar experiments for time-series regression and continuous normalizing flows using neural ODEs. 
	
	Common challenges in both \otd{} and \dto{} are the computational costs associated with solving the ODE  during forward propagation and, when using gradient-based optimization, the calculation of the gradient via backpropagation.
	Accurately solving the forward propagation requires vast quantities of memory and floating point operations, which can make training prohibitively expensive.
	As data in scientific machine learning applications is often noisy, using a low-accuracy solver to speed up computations appears tempting. However, when the forward propagation and adjoint are not solved accurately, the quality of the gradient can deteriorate, an important drawback of the \otd{} approach~\citep{gholami2019anode}.
	In \dto{}, the accuracy of gradients does not depend on the accuracy of the forward propagation.
	This crucial advantage of \dto{} motivates its use to improve the efficiency of training; \anode{} provide detailed analysis and examples on image classification tasks.
	\dto{} approaches include the original formulations of ResNets~\citep{he2016deep} as well as the first continuous ResNets~\citep{weinan2017,HaberRuthotto2017}.
	Finally, \dto{} approaches are easy to implement, especially when using automatic differentiation~\citep{nocedal2006numerical}, while \otd{} methods require the numerical solution of the adjoint equation~\citep{bliss1919}.
	
	Both approaches require storing or recomputing intermediate hidden features from the forward propagation to differentiate the loss function wit respect to the neural network's weights during gradient-based optimization. 
	While \citet{chen2018neural} propose to avoid these memory costs by recomputing the neural ODE backward in time during the adjoint solve, time reversal of ODEs is known to be prone to numerical instabilities~\citep[see example in][]{gholami2019anode}. 
	Avoiding the storage altogether is possible when using forward-backward stable networks, e.g., the Hamiltonian network~\citep{ChangEtAl2017Reversible,RuthottoHaber2019}, but not necessarily for any architecture. For general network structures, checkpointing schemes can be used to reduce the memory footprint without loss of stability~\citep{gholami2019anode}.
	
	In this paper, we compare the \otd{} and \dto{} neural ODE applications in time-series regression and continuous normalizing flows (CNFs).
	Here, meaningful models must also satisfy additional requirements, e.g., the invertibility of the CNF.
	As the continuous model satisfies these requirements by design, Opt-Disc approaches may appear advantageous.
	However, due to the prevalence, importance, and interest in neural ODEs for scientific applications, we believe that a comparison of \otd{} and \dto{} comparison is warranted.
	Through our numerical experiments, we demonstrate that with careful numerical treatment, \dto{} methods can  achieve competitive performance to Opt-Disc at inference with drastically reduced training costs.\footnote{Open-source code is provided at \url{https://github.com/EmoryMLIP/DOvsOD_NeuralODEs}.}
	To this end, we decouple the weights and layers, which allows us to train the neural ODE with a reduced number of layers and satisfy the additional requirements through re-discretization.
	For this approach to be effective, we highlight the importance of using sufficiently small step sizes in the \dto{} approach.
	We show that a too coarse discretization can compromise the invertibility.
	Therefore, our results align with and extend the image classification results achieved in \anode{} to these learning problems using dynamical systems.

	\begin{figure*}
	  	\centering
			\includegraphics[clip, trim=0cm 0cm 0cm 1cm,width=4in]{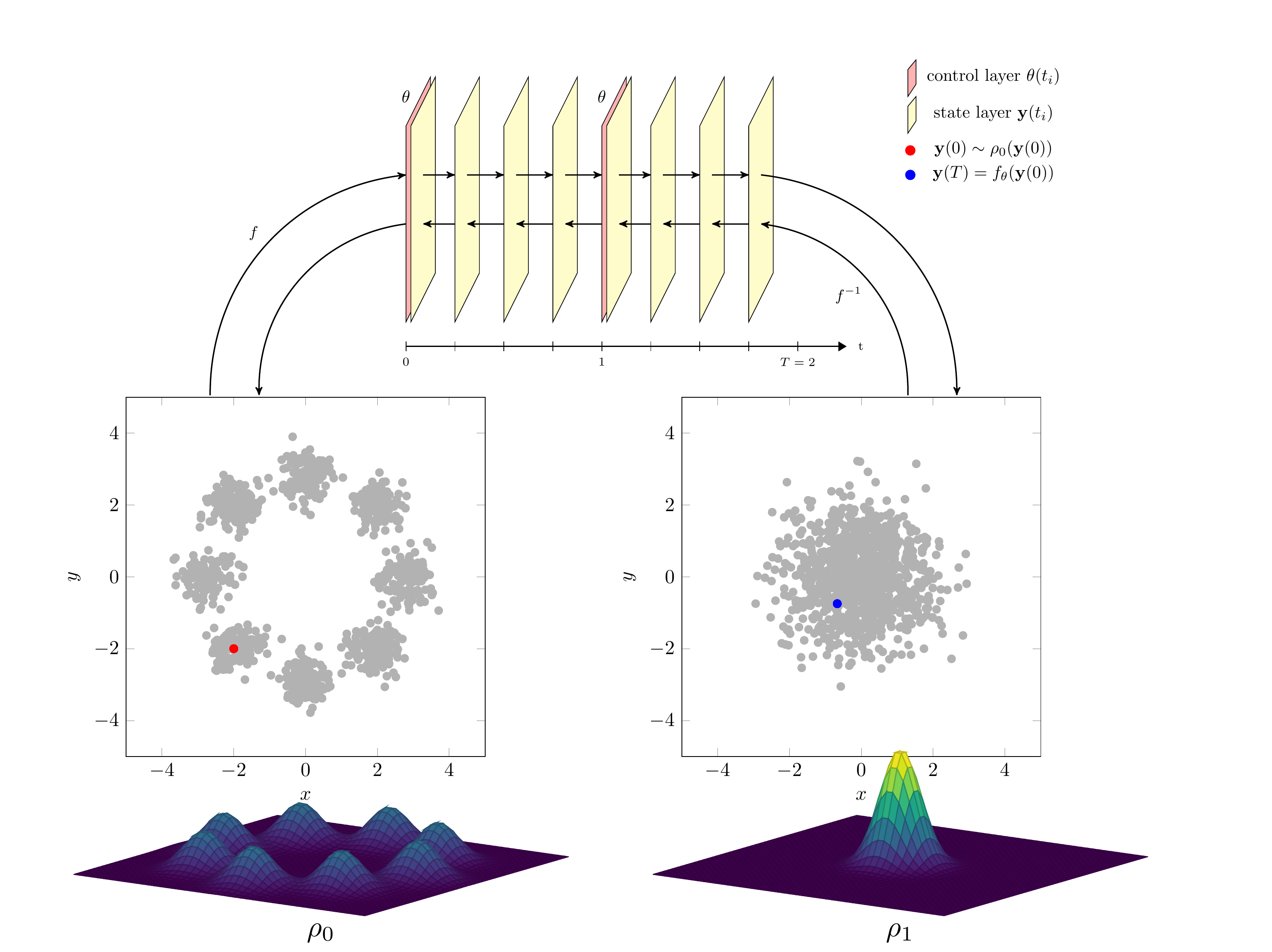}\\
			\caption{Illustration of the \dto{} approach for the Gaussian mixture problem in Section~\ref{sec:sub_8gauss}. As common in optimal control, we use different discretization points for the weights and the states.
			Following the terminology in optimal control, we refer to the time points as  control and state layers, respectively. 
			We discretize the state equation using a Runge-Kutta 4 scheme with constant step size $h=1$. Image is not to scale.}
			\label{fig:cnf}
	\end{figure*}

\section{Background} \label{sec:background}

	We define neural ODEs, place them in the context of discrete neural networks, and provide a brief groundwork of optimal control theory for the \dto{} and \otd{} comparison.

	\subsection{Neural ODEs}
	
	Neural ODEs, coined by \citet{chen2018neural}, are differential equations with neural network components.
	The neural ODE performs a nonlinear transformation of the feature space. 
	We denote the transformation of a feature vector $\bfy_0 \in \mathbb{R}^{n_f}$ by model $f(\bfth,\bfy_0) = \bfy(T)$ where $T$ is some arbitrary final time and $\bfy \colon [0,T] \to \mathbb{R}^{n_f}$ satisfies the initial value problem
	\begin{equation} \label{eq:cont_ode}
		\begin{aligned}
			\partial_t \bfy(t) &= \ell(\bfth(t), \bfy(t), t ) \, , \quad \text{for} \quad t \in (0, T] \\
			\bfy(0) &= \bfy_{0}.
		\end{aligned}
	\end{equation}
	Here, $\bfth \colon [0,T] \rightarrow \mathbb{R}^{n_p}$ are the weights of the neural ODE, and $\ell \colon \mathbb{R}^{n_p} \times \mathbb{R}^{n_f} \times [0,T] \to \mathbb{R}^{n_f}$ is a neural network. We formulate layer $\ell$ to accept time $t$ as a parameter though only in rare cases (Section~\ref{sec:cnf}) does a neural network layer operate on time itself.

	A prominent class of neural ODEs are continuous limits of ResNets~\citep{he2016deep}. 
	An  $N$-layer ResNet can be written as
	\begin{equation} \label{eq:f_euler}
		\bfy_{j+1} = \bfy_{j} + h \, \ell\left( \bfth_j , \bfy_{j}, t_j \right), \quad \text{where} \quad j=0,1,\ldots,N{-}1,
	\end{equation}
	with some function $\ell$ and step size $h = T/N$. We refer to the $\bfth_j$ as \textit{control layers} and the $\bfy_{j}$ as \textit{state layers} (Figure~\ref{fig:cnf}).
	Interpreting the weights $\bfth_j$ as evaluations of $\bfth$ at the time points $t_j = j \cdot h$ and taking the limit as $N\to\infty$, we see that the $N$-layer ResNet~\eqref{eq:f_euler} converges to the neural ODE~\eqref{eq:cont_ode}. Infinitely deep networks such as this have outperformed shallower networks in recent years~\citep{sonoda2019transport}.
	Analogously, one can view the $N$-layer ResNet as a forward Euler discretization of the neural ODE~\citep{weinan2017,HaberRuthotto2017}.
	This observation led to follow-up works on analyzing and improving ResNets via insights from the continuous model.
	Since ResNets are discrete processes and ultimately a discrete network is desired, some caution must be used when studying their continuous limits~\citep{ascher2019discrete}.
	
	In this paper, we consider a different class of neural ODEs arising in time-series regression and CNFs. 
	Unlike ResNets, this class of learning problems is phrased in the continuous setting, and training aims at obtaining the continuous neural ODE model for prediction or inference. 
	These classes also impose additional requirements on meaningful models, e.g., the ability to reverse the neural ODE in time.
	While the continuous neural ODE satisfies these requirements by design, discrete versions may not.

	\subsection{Learning and Optimal Control}

	Training the neural ODE~\eqref{eq:cont_ode} can be phrased as an infinite-dimensional optimal control problem
	\begin{equation} \label{eq:opt}
		\begin{aligned}
		& \min_{\bfth,\bfy}
		& &  \left\{ \calJ(\bfth,\bfy) :=  \frac{1}{S} \sum_{i=1}^{S}\calL(\bfth,\bfy^{(i)}) + \calR(\bfth)\right\}\\
		& \text{subject to} & & \bfy^{(i)} \text{ solves }\text{\eqref{eq:cont_ode}} \text{ with initial value } \bfy_0^{(i)} \\
		\end{aligned}
	\end{equation}
	where $\bfy^{(1)}, \ldots, \bfy^{(S)}$ satisfy the neural ODEs for initial values given by the training data; loss functional $\calL$ and regularization functional $\calR$ are chosen to model a given learning task.
	For regularization, we consider Tikhonov regularization (often called weight decay,~\citealp{krogh1992weightdecay})
	 \begin{equation}
	 	\calR(\bfth) = \frac{\alpha}{2} \int_{0}^T \| \bfth(t) \|^2 \, \du t,
	 \end{equation}
	 where we assume that the regularization parameter, $\alpha>0$, is chosen and kept fixed.
	 
	 We will consider two types of loss functionals; regression loss and likelihood. The former is used when training a neural ODE to approximate a given function $\bfu \colon [0,T] \to \mathbb{R}^{n_f}$ and reads
	\begin{equation}\label{eq:Lreg}
		\calL(\bfth,\bfy) = \int_{0}^T L\big(\bfy(t),\bfu(t)\big) \, \du t, \quad L(\bfy,\bfu) =  \frac12 \| \bfy - \bfu \|^2 .
	\end{equation}
	When the $\bfu$ is known only at some time points, a discretized version of this functional $\calL$ is used (e.g., the time-series example in Section~\ref{sec:times_series}). 
	Describing the likelihood function used in CNFs is more involved, and we postpone its discussion to Section~\ref{sec:cnf}.

  	Existing training approaches for neural ODEs almost exclusively consider the reduced version of~\eqref{eq:opt}, which is obtained by solving the ODE~\eqref{eq:cont_ode} for fixed weights. 
	Denoting the unique solution of the state equation~\eqref{eq:cont_ode} for weights $\bfth$ by $\bfy(\bfth)$, we obtain the reduced optimization problem 
	\begin{equation}\label{eq:reduced_opt}
		\min_{\bfth} \quad  \calJ(\bfth,\bfy(\bfth)), 
	\end{equation}
	which is an infinite-dimensional unconstrained problem in the controls $\bfth \colon [0,T] \to \mathbb{R}^{n_p}$. 
	
	To transform the variational problem~\eqref{eq:reduced_opt} into a finite-dimensional problem, it is most common to  discretize the control on a control grid (e.g., $\bfth$ at $t=0,1$ in Figure~\ref{fig:cnf}) and numerically approximate the integrals in $\calJ$ using quadrature rules. 
	We use this strategy and in Section~\ref{sec:DOOD} focus on handling the state equation~\eqref{eq:cont_ode}. 
	To this end, two main approaches exist; the optimize-discretize (\otd{}) approach optimizes the differential equation and discretizes the ODE only for the optimal weights, while the discretize-optimize (\dto{}) approach uses a numerical time integrator to obtain a fully discrete version of~\eqref{eq:reduced_opt} that is then solved to determine the optimal weights.

\section{Discretize-Optimize and Optimize-Discretize}\label{sec:DOOD}

	The \dto{} approach to performing time integration discretizes the ODE~\eqref{eq:cont_ode} in time and then optimizes over the discretization. 
	This approach is common in neural networks and easy to implement, especially when automatic differentiation can be used. 
	For example, a discrete ResNet with input features (or initial conditions) $\bfy_0$ follows forward propagation~\eqref{eq:f_euler} as the explicit Euler method on a uniform time discretization with step size $h$.
	The choice of fixed step size $h$ trades off the accuracy of the solution with the amount of computation and overall training time.

	In applications that require the neural ODE for prediction or inference, such as time-series regression and CNFs, the step size must be chosen judiciously to ensure the trained discrete model captures the properties of the continuous model.
	Choosing a too large value of $h$, may, for example, yield a discrete flow model with an inaccurate or suboptimal inverse.
	Choosing a too small value of $h$ leads to greater computational costs.
	In contrast to selection of other training hyperparameters, the step size choice provides the advantage of monitoring the accuracy of the ODE solver.
	While we use a fixed step size $h$ for simplicity, we can obtain even more efficient approaches by combining adaptive discretization schemes with the backpropagation used in \dto{}. 

	Following similar steps used in \anode{}, we expose a crucial difference in the gradient computation between the \dto{} and \otd{} approaches.
	For ease of presentation, we use and assume that the forward propagation is the forward Euler scheme~\eqref{eq:f_euler} and that the objective function consists of the time-series regression loss defined by~\eqref{eq:Lreg} with no regularization.
	Accurate gradients are critical in ensuring the efficiency of gradient-based optimization algorithms, including Stochastic Gradient Descent (SGD)~\citep{robbins1951} and ADAM~\citep{kingma2014adam}.
	The updates for $\bfth$ in such methods depend on the gradient of the objective function in~\eqref{eq:reduced_opt}. 
	
	In \dto{}, the backpropagation of the discrete ResNet~\eqref{eq:f_euler} computes the gradients. 
	Automatic differentiation, which traverses the computational graph backward in time, is used commonly in machine learning frameworks.
	The discretization of the forward propagation completely determines this process. For discrete objective function
	\begin{equation}
		J(\bfth) = h \sum_{i=1}^N L(\bfy_i,\bfu_i)
	\end{equation}
	and using auxiliary variable $\bfz$, the backpropagation through the forward Euler discretization in~\eqref{eq:f_euler} is
	\begin{equation} \label{eq:back_do}
		\begin{split}
		\nabla_{\bfth_j} J(\bfth)  &= h \nabla_{\bfth} \, \ell(\bfth_j,\bfy_j,t_j) \, \, \bfz_j , \quad \text{where} \\
		\bfz_{N} &= h \, \nabla_{\bfy}\regL(\bfy_N, \bfu_N) \quad \text{and} \\
		\bfz_{j}   &= \bfz_{j+1} + h \left( \nabla_{\bfy}\ell(\bfth_j,\bfy_j, t_j) \bfz_{j+1} + \nabla_{\bfy}\regL(\bfy_j, \bfu_j) \right),  \\
		\end{split}
	\end{equation}
	and the computations are backward through the layers, i.e., $j=N{-}1, N{-}2, \ldots, 1$.

	In \otd{} the gradients are computed by numerically solving the adjoint equation
	\begin{equation} \label{eq:back_cont}
	 \begin{split}
		\nabla_{\bfth(t)} \calJ(\bfth)  &= \bfz(t) \nabla_{\bfth} \ell \big(\bfth(t),\bfy(t),t\big), \quad \text{where}\\
		\bfz(T) &= \nabla_{\bfy} L\big(\bfy(T),\bfu(T)\big) \quad \text{and} \\
		- \partial_t \bfz &=   \nabla_{\bfy}\ell\big(\bfth(t),\bfy(t),t\big)  \bfz(t) + \nabla_{\bfy} L\big(\bfy(t),\bfu(t)\big). \\
	 \end{split}
	\end{equation}
	As indicated by the notation $- \partial_t $, this final value problem is solved backward in time.
	This equation can be derived from the Karush-Kuhn-Tucker (KKT) optimality conditions of the continuous learning problem (Appendix~\ref{app:adj_eqns}).
			
	Comparison of the backpropagation~\eqref{eq:back_do} and the adjoint computation~\eqref{eq:back_cont} 
	shows that both depend on the intermediate states, which need to be stored or recomputed.
	However, the two differ because flexibility exists for choosing the numerical scheme to discretize the adjoint computation in~\eqref{eq:back_cont}, whereas the computation in~\eqref{eq:back_do} is determined by the discrete forward propagation.
	In fact, the backpropagation shown in~\eqref{eq:back_do} can be seen as a discretization of the adjoint equation; however, the standard backward Euler scheme reads
	\begin{equation} \label{eq:back_od}
		\begin{split}
		\nabla_{\bfth_j} J(\bfth)  &= h \nabla_{\bfth} \ell \big(\bfth(t_j),\bfy(t_j),t_j\big) \bfz_j, \quad \text{where} \\
		\bfz_{N} &= h \nabla_{\bfy}\regL\big(\bfy(t_N), \bfu(t_N)\big) \quad \text{and} \\
		\bfz_{j}   &= \bfz_{j+1} + h \big(  \nabla_{\bfy}\ell \big( \bfth(t_{j+1}),\bfy(t_{j+1}), t_{j+1} \big) \bfz_{j+1} + \nabla_{\bfy}\regL \big(\bfy(t_{j+1}), \bfu(t_{j+1}) \big) \big). \\
		\end{split}
	\end{equation}
	The only difference between~\eqref{eq:back_do} and~\eqref{eq:back_od} is a shift of the indices of the intermediate weights and features.
	Thus, the gradients obtained using both methods differ unless both equations are solved accurately, such as when $h$ converges to zero. 
	Hence, the \otd{} approach in~\citet{chen2018neural}, which uses adaptive time integrators for the forward and adjoint equations, may provide inaccurate gradients when the time steps differ between both solvers and the tolerance of those solvers is not sufficiently small. 
	This comparison is fairly standard and is also performed in \anode{}.

	The differences of the gradients computed using \dto{} and \otd{} affect the convergence of neural ODEs for image classification~\citep{gholami2019anode}. 
	Since the gradient in the \dto{} method is an accurate estimate of the gradient of the discrete objective function for any accuracy, the differences are most striking when solving the forward propagation inaccurately.
	Data science and scientific applications may desire an inaccurate forward propagation since the data may be far less accurate than the tolerance setting necessary for rendering a useful gradient in the \otd{} approach.
	In the context of time-series regression and continuous normalizing flows, the potential savings are limited since the discretization trained in \dto{} must capture the relevant properties of the continuous model.

\section{Numerical Experiments} \label{sec:ne}

We compare the \dto{} approach with the \otd{} approach for two use cases: time-series regression and  density estimation through continuous normalizing flows. 
For the former, we use the test data and experimental setup of the \otd{} in~\citet{rackauckasBlogPost} to obtain a direct comparison.
For the latter, we use a synthetic data set and five sizable public data sets using CNFs as in~\citet{grathwohl2018ffjord}.

\subsection{Time-Series Regression} \label{sec:times_series}

	\begin{figure*}
	   	\centering
	 		\includegraphics[width=0.3\linewidth]{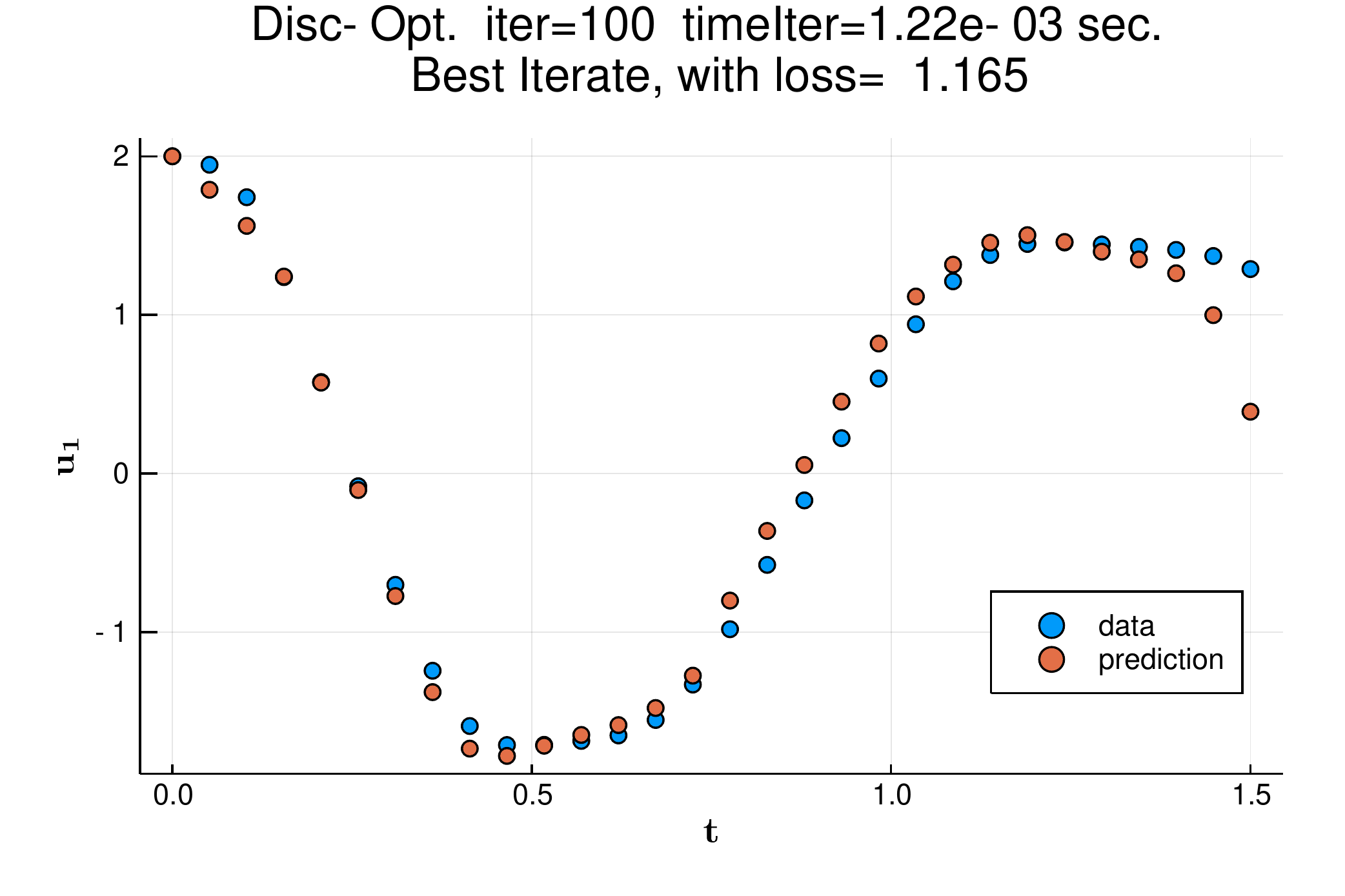}%
	 		\includegraphics[width=0.3\linewidth]{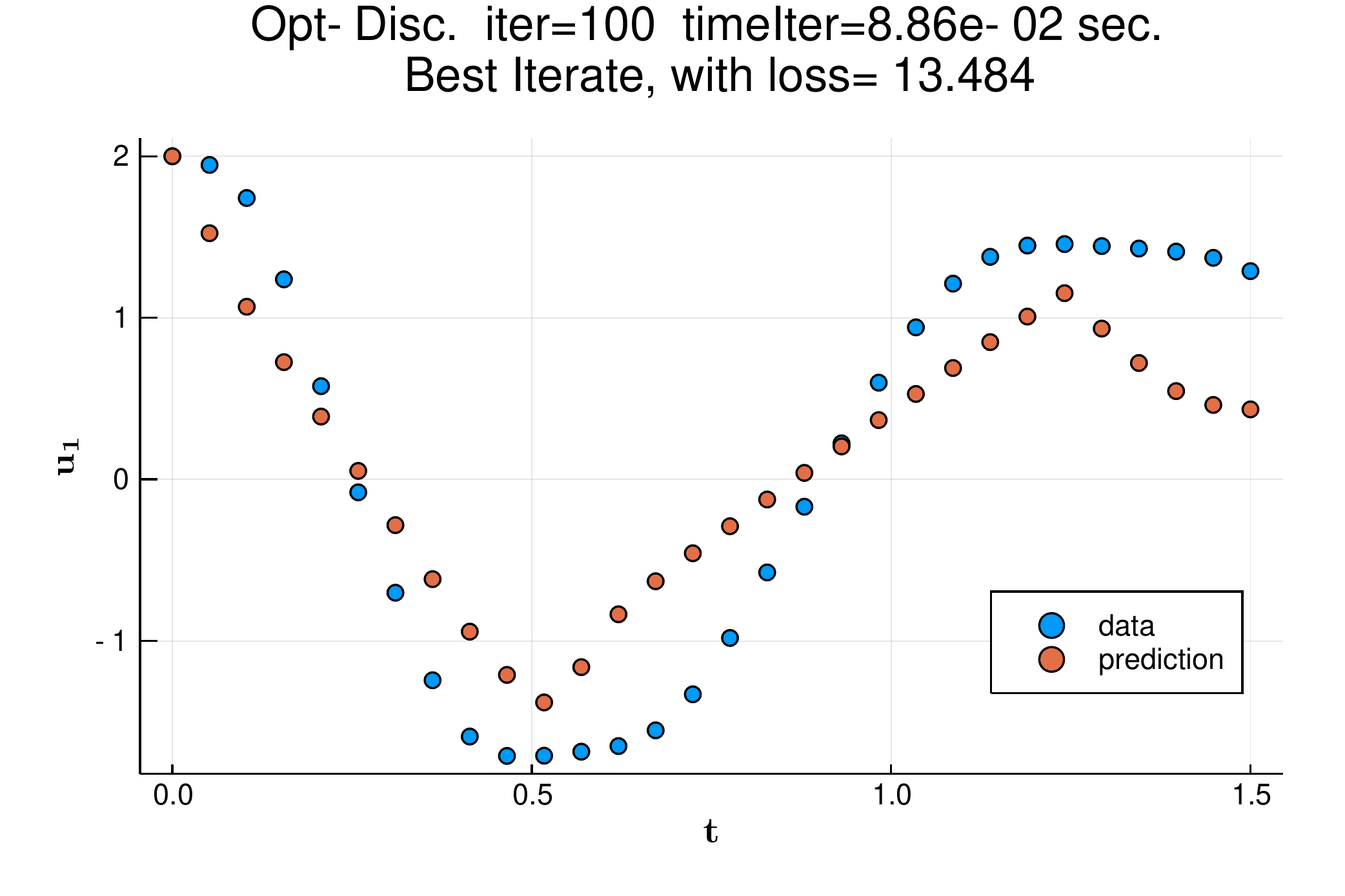}\\
	 		\includegraphics[width=0.3\linewidth]{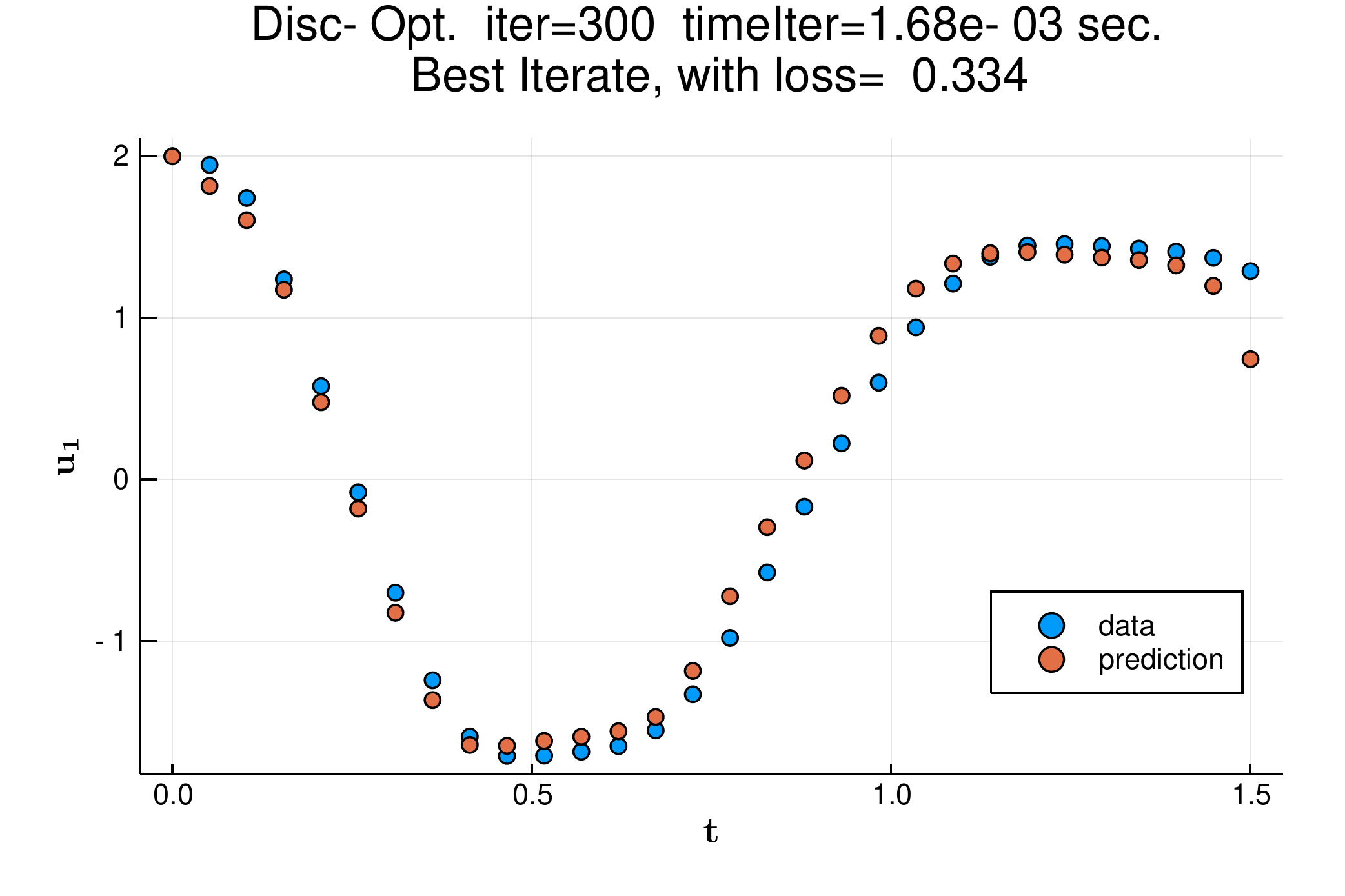}%
	 		\includegraphics[width=0.3\linewidth]{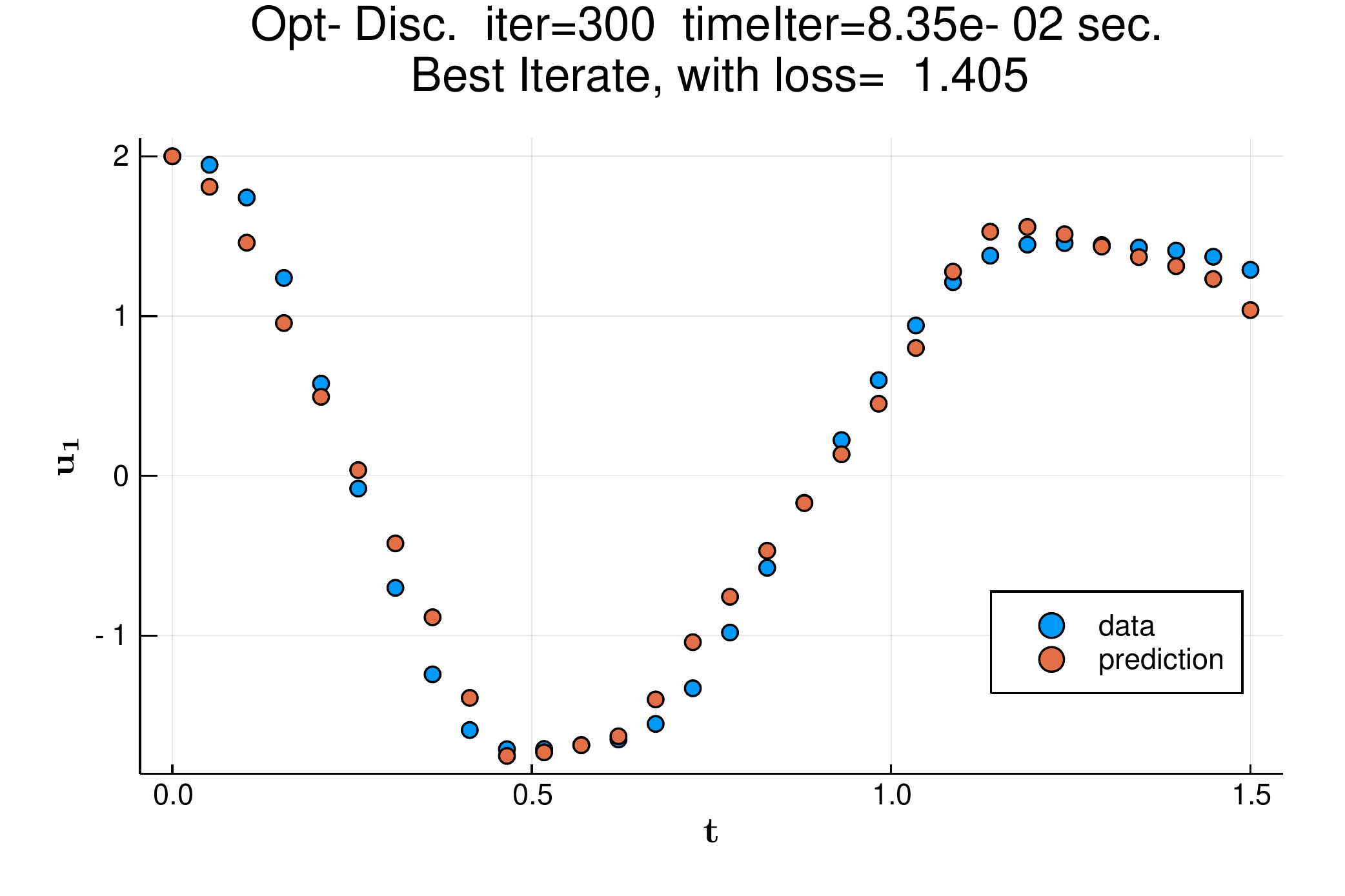}\\
	 		\caption{Time-series regression training iterations 100 and 300 comparing the \dto{} and \otd{} approaches with the ground truth~\eqref{eq:juliaODE}. Comparative convergence video of the two methods is available at \url{https://imgur.com/nWxwVoe}. 
	 		Discrepancies between convergence behavior of the approaches vary with initial parameterization (Section~\ref{sec:seeds}). }
	 		\label{fig:juliaCollage}
	\end{figure*}

	We consider the time-series regression problem in \citet{chen2018neural} and~\citet{rackauckasBlogPost} to perform a direct comparison between \otd{} and \dto{}. 
	Given time-series data  $\bfu_1=\bfu(t_1), \ldots,\bfu_n=\bfu(t_n)$ obtained from some unknown function $\bfu$, the goal is to tune the neural ODE weights $\bfth$ in \eqref{eq:cont_ode} such that $\bfy(t_k) \approx \bfu_k$ for $k=0, \ldots ,n$.
	The initial value of the  neural ODE is $\bfy(0) = \bfy_0 = \bfu_0$. 
 	
	We use the data set generated in~\citet{rackauckasBlogPost}, where the data is obtained from the  ODE model
	\begin{equation} \label{eq:juliaODE}
		\begin{cases}
			& \partial_t \bfu = \bfA \bfu^{\circ 3} 						  \\
			& \bfu_0 = 
			\left[\def\arraystretch{0.9}
			\begin{array}{c} 2 \\ 0 \end{array} \right] \\
		\end{cases}
		\quad , \qquad \text{where \,} \bfA = \left[ \begin{array}{cc} -0.1 & 2 \\ -2 & -0.1 \end{array} \right]
	\end{equation}
	where $\bfu^{\circ 3}$ denotes the element-wise cubic and $t \in [0,1.5]$.
	Using the adaptive Tsitouras 5/4 Runge-Kutta method (dopri5), the ODE is solved at 30 equidistantly-spaced time points (Figure~\ref{fig:juliaCollage}).
	As a loss function, we use the regression loss in~\eqref{eq:Lreg}.

	We use the same neural ODE as in~\citet{rackauckasBlogPost} that is based on the layer
	\begin{equation}
			\ell(\bfth,\bfy_0,t) \, \colonEquals \, \myD{2} \circ \tanh \circ \, \myD{1}(\bfy_0^{\circ 3}).
	\end{equation}
	Here, $\myD{i}$ denotes a single linear layer that performs an affine transformation of the inputs and is parameterized by $\bfth_i$ for $i=1,2$. 
	The neural ODE is trained using 300 steps of the ADAM optimizer with step size $0.1$, starting with a Glorot initialization~\citep{glorot2010understanding}. The \otd{} and \dto{} approaches use the same initializations (i.e., same random seed). 
	
	The \otd{} approach in~\citet{rackauckasBlogPost}  uses the default dopri5 solver.
	We compare this with a \dto{} approach that uses \rk{} with a fixed step size $h$ for training the neural ODE. For the step size $h$, we use the spacing of the data points, i.e., $h=\frac{1.5}{29}$. With this choice, the evaluation of the loss function does not require any interpolation. 
	
\begin{figure*}
	  	\centering
			\includegraphics[clip, trim=.1cm .1cm .1cm .1cm,width=.32\linewidth]{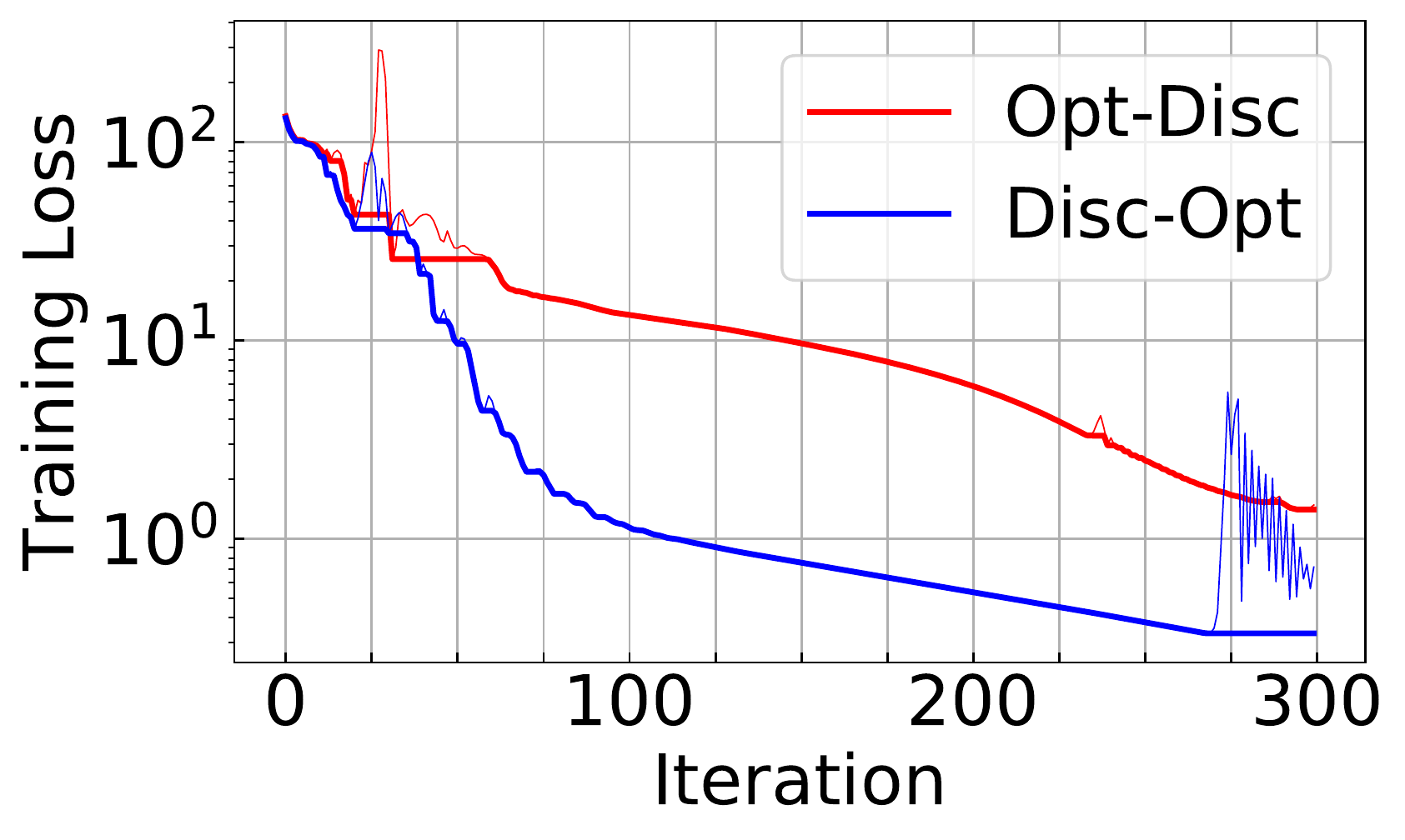}
			\includegraphics[clip, trim=1cm .1cm .1cm .1cm,width=.3\linewidth]{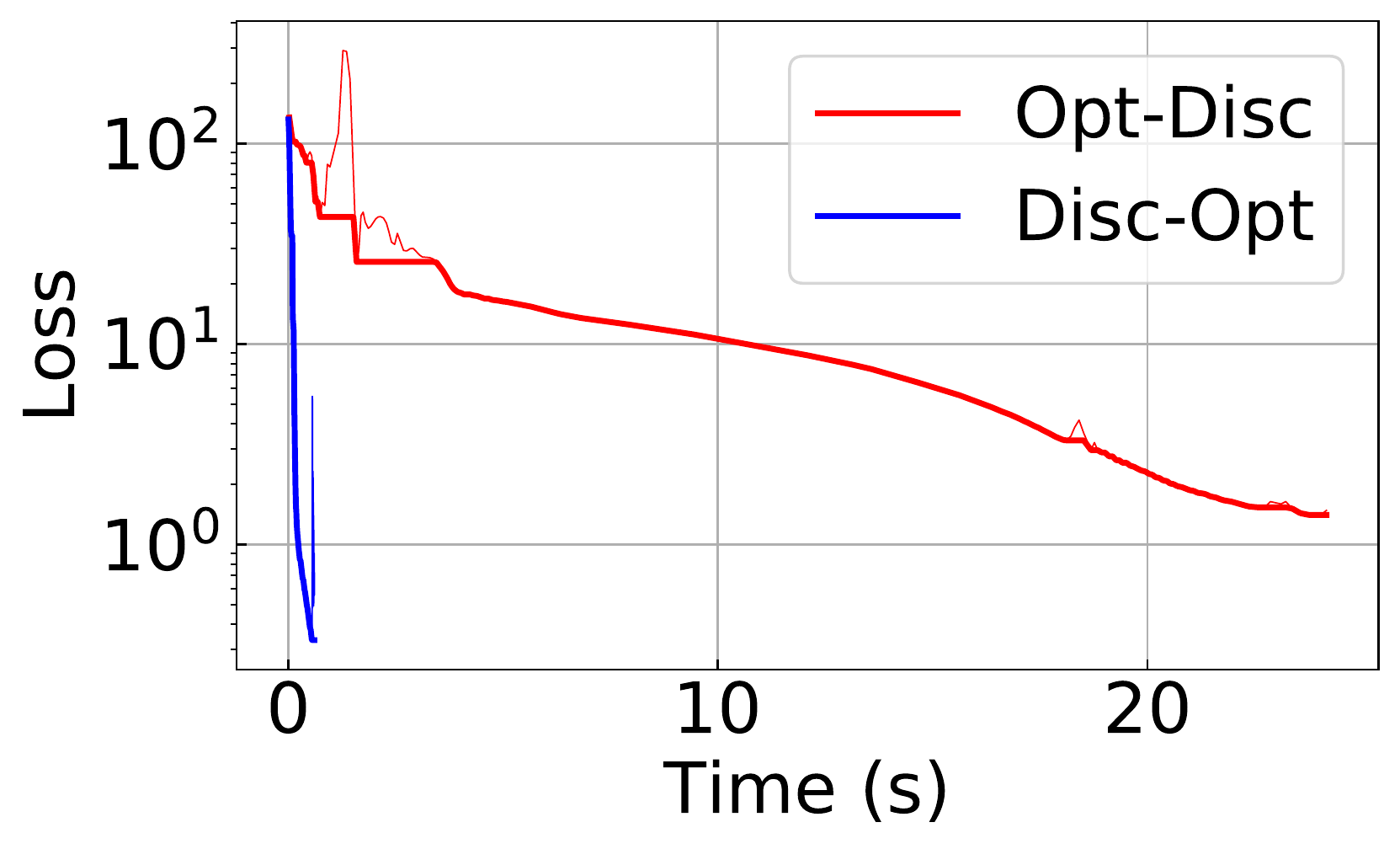}
			\caption{For time-series regression, the \dto{} method converges in fewer iterations, and each of its iterations requires less time. The mean iteration clocktimes are 2.0 ms for \dto{} and 80.5 ms for \otd{}. Around iteration 24, the ODE solvers struggle to improve the training (cf. Figure~\ref{fig:derivCheck}), but the \otd{} approach appears to suffer more.}
			\label{fig:timeAndLoss}
	\end{figure*}

	The \dto{} approach substantially reduces the cost of training. We attribute these savings to the following two reasons.
	First, the runtime per iteration is 97\% lower than the \otd{} approach's runtime per iteration (Figure~\ref{fig:timeAndLoss}). 
	This speedup mainly comes from the fewer function evaluations, pre-determined by the step size choice, used in the \dto{} approach.
	Since the  \otd{} method uses adaptive time-stepping, the number of function evaluations, and thus, cost per iteration varies during the optimization.
	Second, the \dto{} approach also converges in roughly one third of the total iterations (Figure~\ref{fig:timeAndLoss}) with predictions at iteration 100 drastically closer to the ground truth than the \otd{} approach (Figure~\ref{fig:juliaCollage}).
	
\begin{figure*}
	  	\centering
			\includegraphics[width=.32\linewidth]{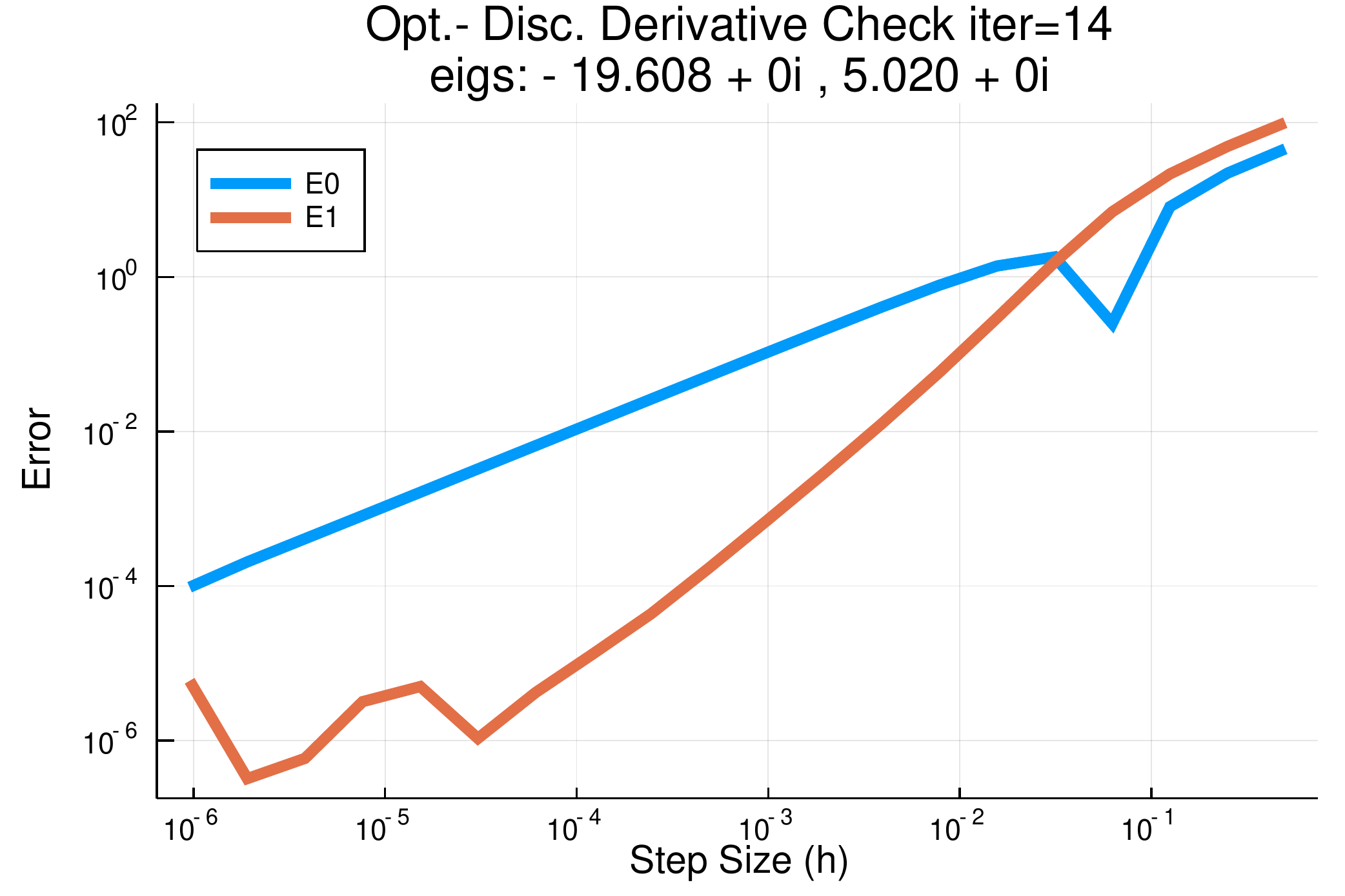}
			\includegraphics[width=.32\linewidth]{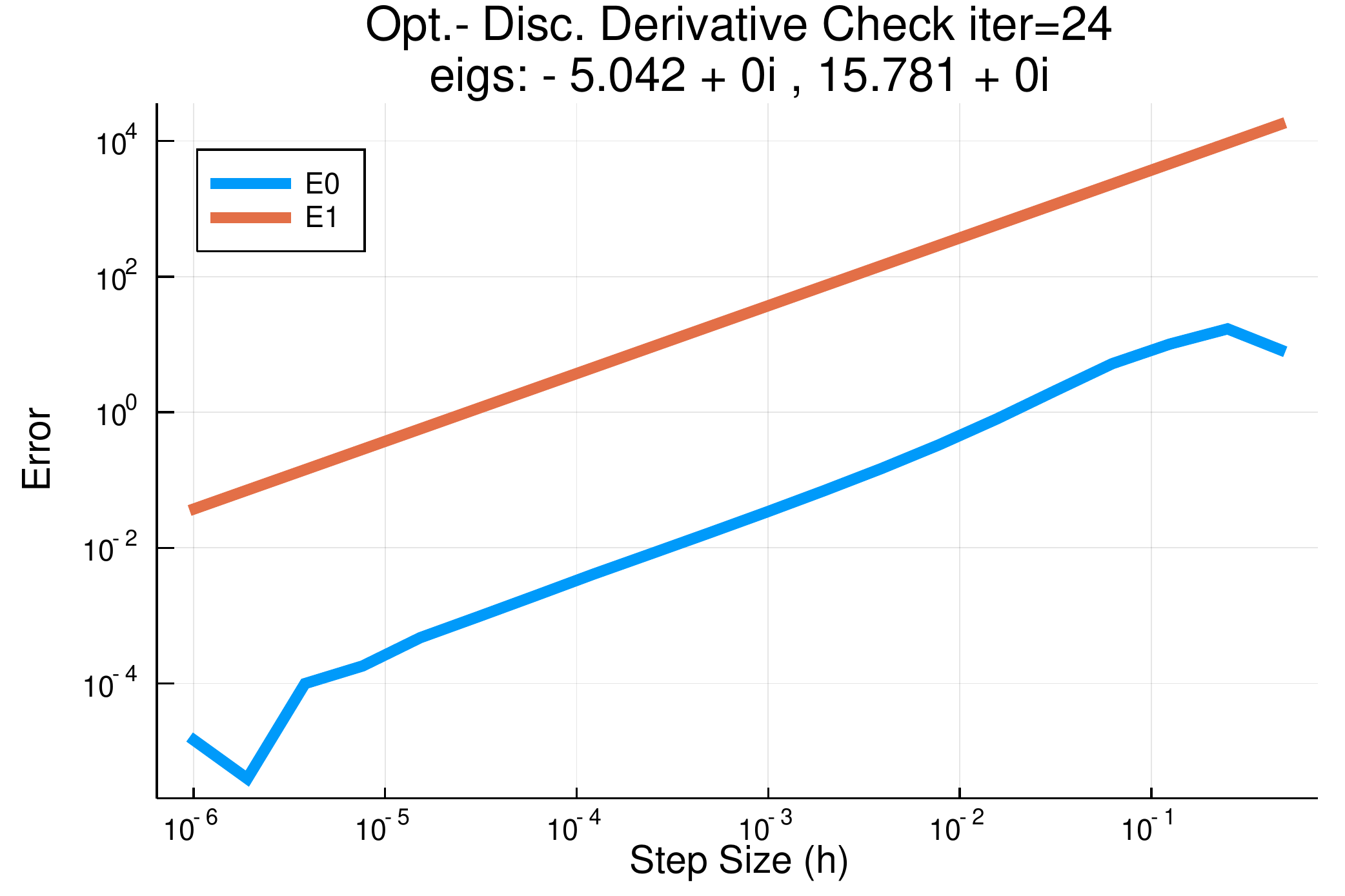}
			\\
			\includegraphics[width=.32\linewidth]{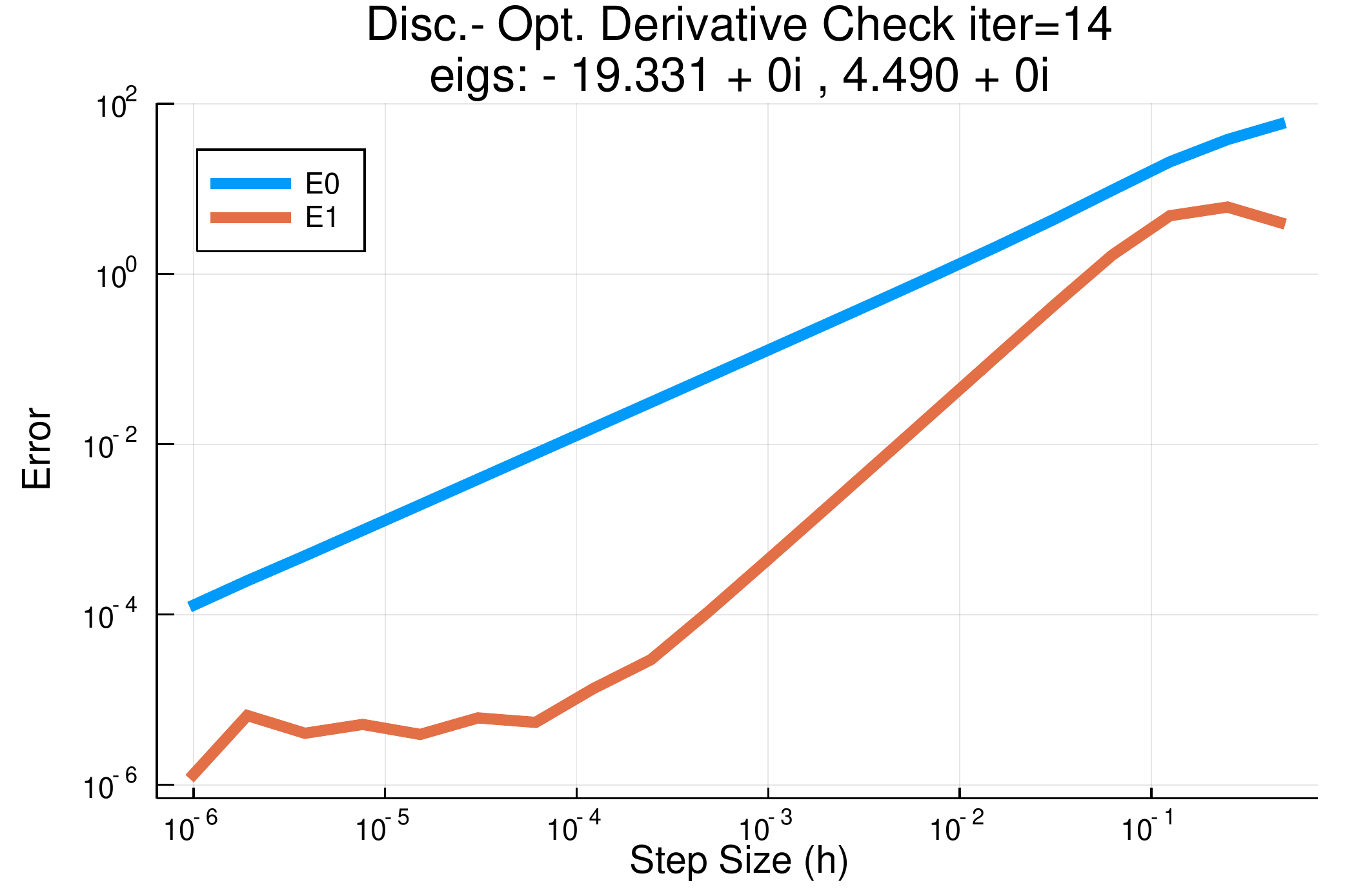}
			\includegraphics[width=.32	\linewidth]{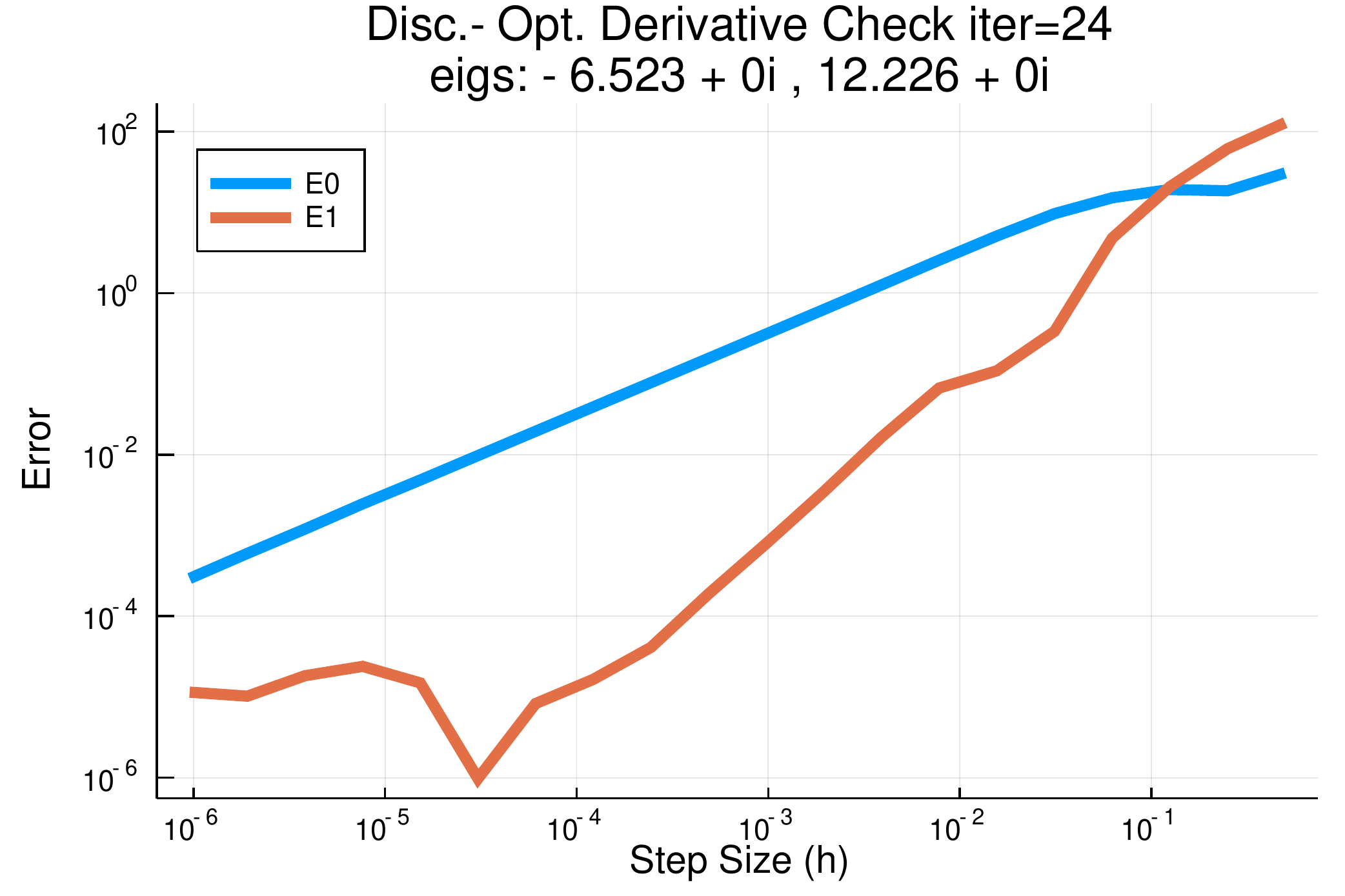}
			\caption{The derivative check~\eqref{eq:deriv_check} for time-series regression iterations 14 and 24 shown on log-log plot. As expected, the gradients of the \dto{} approach (bottom row) are correct; note the faster decay of $E_1(h)$ (red line) compared to $E_0(h)$ (blue line) as $h$ decays.
			The gradient in the \otd{} approach (top row) is correct for iteration 14, but not for iteration 24. In this case (top right), the function $E_1$ (red line) is greater than $E_0$ (blue line).}
			\label{fig:derivCheck}
	\end{figure*}

	To explain the fewer iterations needed by \dto, we numerically test the quality of the gradients provided by both approaches through use of Taylor's theorem. 
	Let $\bfg \in \mathbb{R}^n$ denote the gradient of the objective function
	 $\bfF \colon \mathbb{R}^n \rightarrow \mathbb{R}$ at a point $\bfth$ 
	  and let $\bfv\in\mathbb{R}^n$ be a randomly chosen direction.
	Then, by Taylor's theorem, we have that
	\begin{equation} \label{eq:deriv_check}
		\begin{aligned}
			E_0(h) &\colonEquals \lVert \bfF(\bfth+h \bfv) - \bfF(\bfth) \rVert = \mathcal{O}(h \|\bfv\|) \quad \text{and} \\
			E_1(h) & \colonEquals \lVert \bfF(\bfth+h \bfv) - \bfF(\bfth) - h \bfg^{\top}\bfv\rVert =  \mathcal{O}(h^2 \|\bfv\|).
		\end{aligned}
	\end{equation} 
	 We observe the decay of $E_0(h)$ and $E_1(h)$ as $h \to 0$ for a fixed (randomly chosen) direction $\bfv$ around the current network weights for two different iterations of the training (Figure~\ref{fig:derivCheck}).
	 We scale both axes logarithmically. 
	 As expected, $E_0$ decays almost perfectly linearly, and the decay of $E_1$ using the \dto{} method is approximately twice as steep for large $h$, ultimately leveling off due to rounding errors and conditioning (Figure~\ref{fig:derivCheck} bottom row). 
	 While the derivative obtained in the \otd{} approach is correct at iteration 14 (Figure~\ref{fig:derivCheck} top left), it fails to provide an accurate gradient at iteration 24 (Figure~\ref{fig:derivCheck} top right), where the error $E_1$ is greater than $E_0$ for all $h$ that we tested.

\subsubsection{Extrapolation and Different Initial Conditions} \label{sec:seeds}

\begin{figure*}
	  	\centering
			\includegraphics[width=0.35\linewidth]{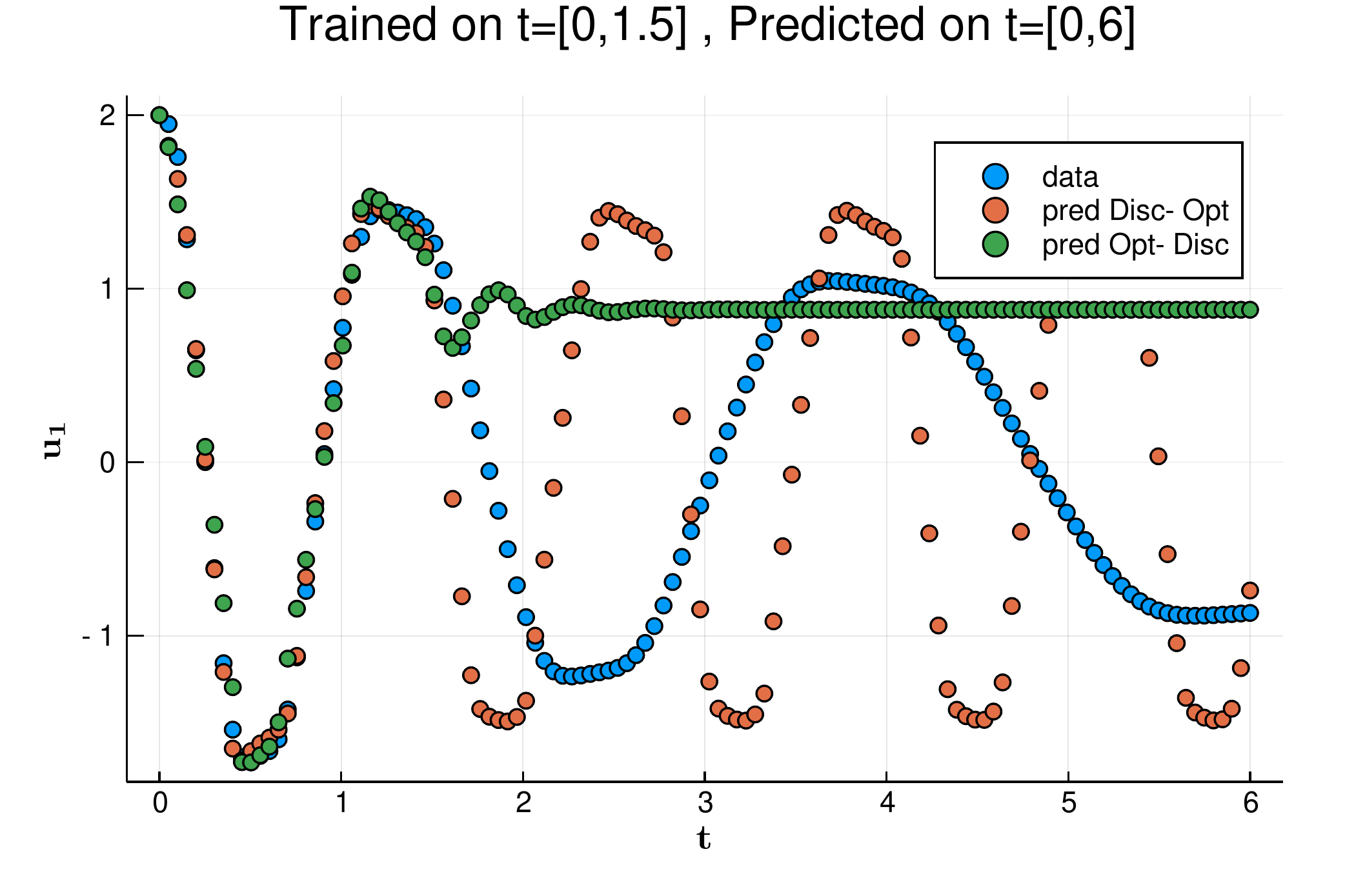}
			\caption{Extrapolation of time-series regression models. After training each model on the time interval $t \in [0,1.5]$, we visualize how the trained neural networks extrapolate up to $t=6$. Each neural ODE models behaves as a smooth ODE, by construction, though the learned models oscillate much more quickly than the ground truth ODE.}
			\label{fig:extrapolation}
	\end{figure*}

	For time-series regression on $t \in [0,1.5]$, the neural networks appear to have modeled the ground truth ODE quite well (Section~\ref{sec:times_series}). We ask if the models extrapolate correctly to time $t$ outside the training period $[0,1.5]$. We extend the prediction of the trained models to the time period $t \in [0,6]$ (Figure~\ref{fig:extrapolation}) in which we see that both the \dto{} and \otd{} approaches learn behaviors that represent smooth ODEs that are different from the ground truth used to create the data. This presents an example where neural ODEs extrapolate poorly.

\begin{figure*}[t]
	\centering
	{%
	\subfloat[Different seed 1]{\label{fig:seed200}%
	    	\includegraphics[width=0.3\linewidth]{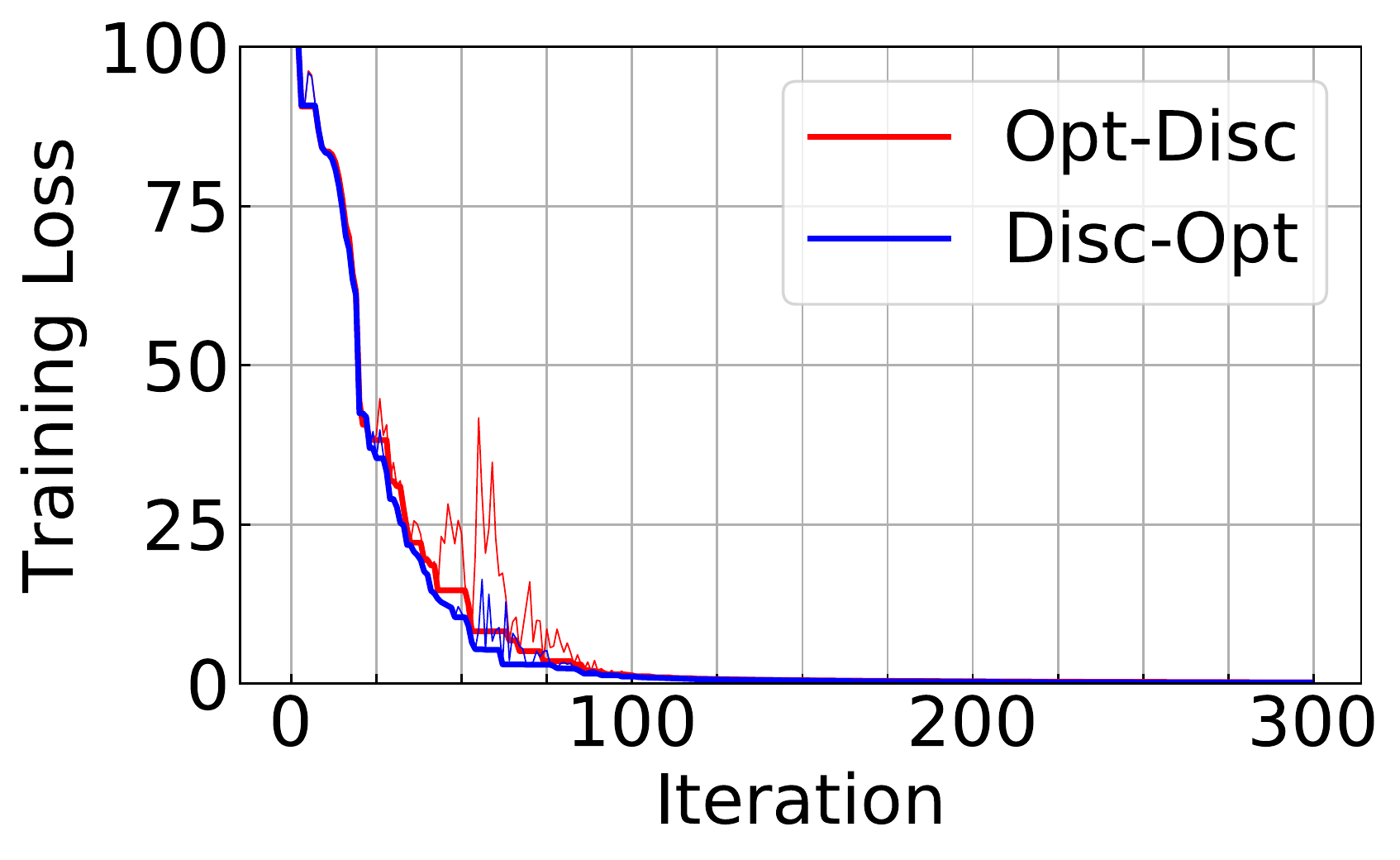}
	      	\includegraphics[width=0.3\linewidth]{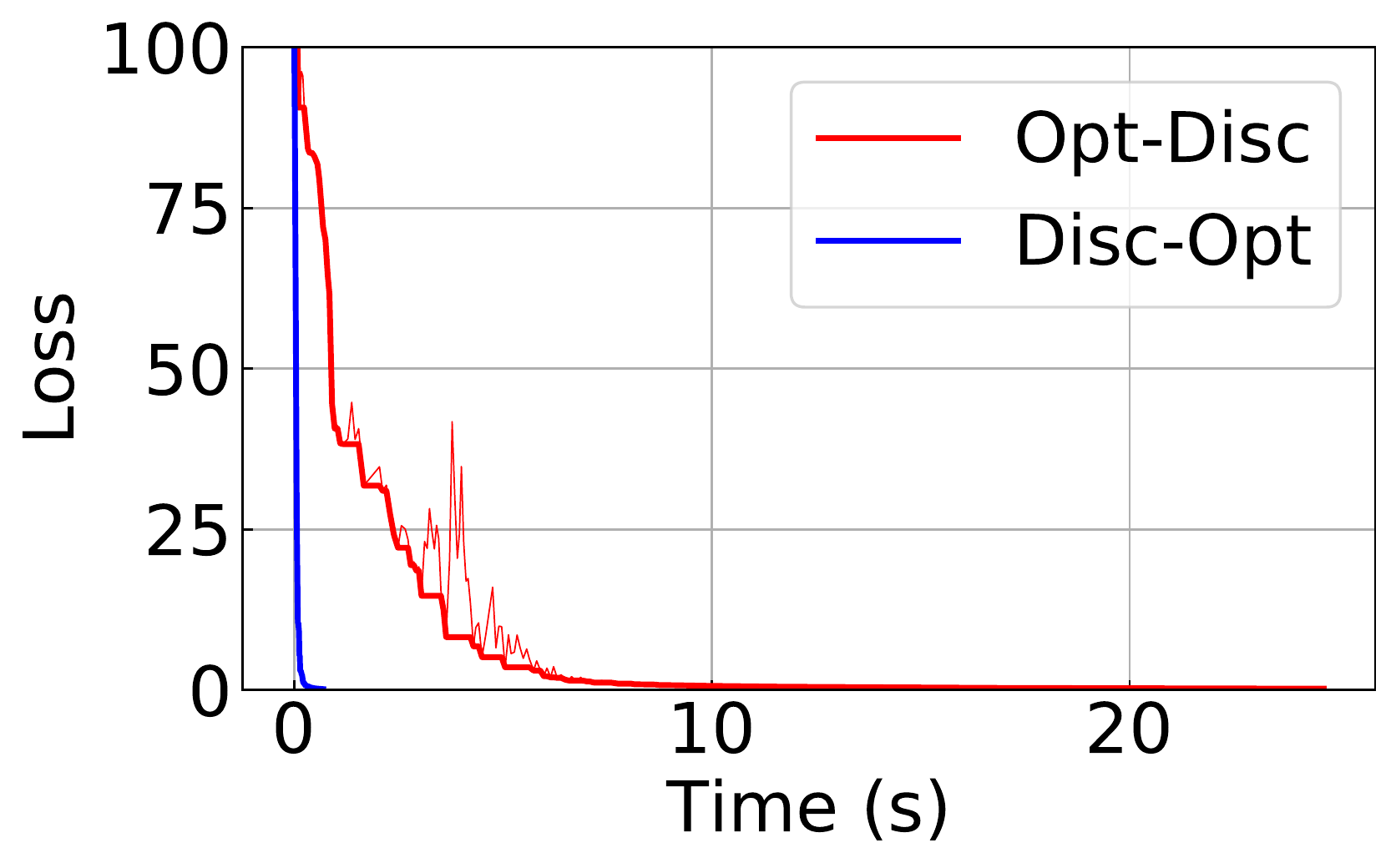}
	      	\includegraphics[width=0.3\linewidth]{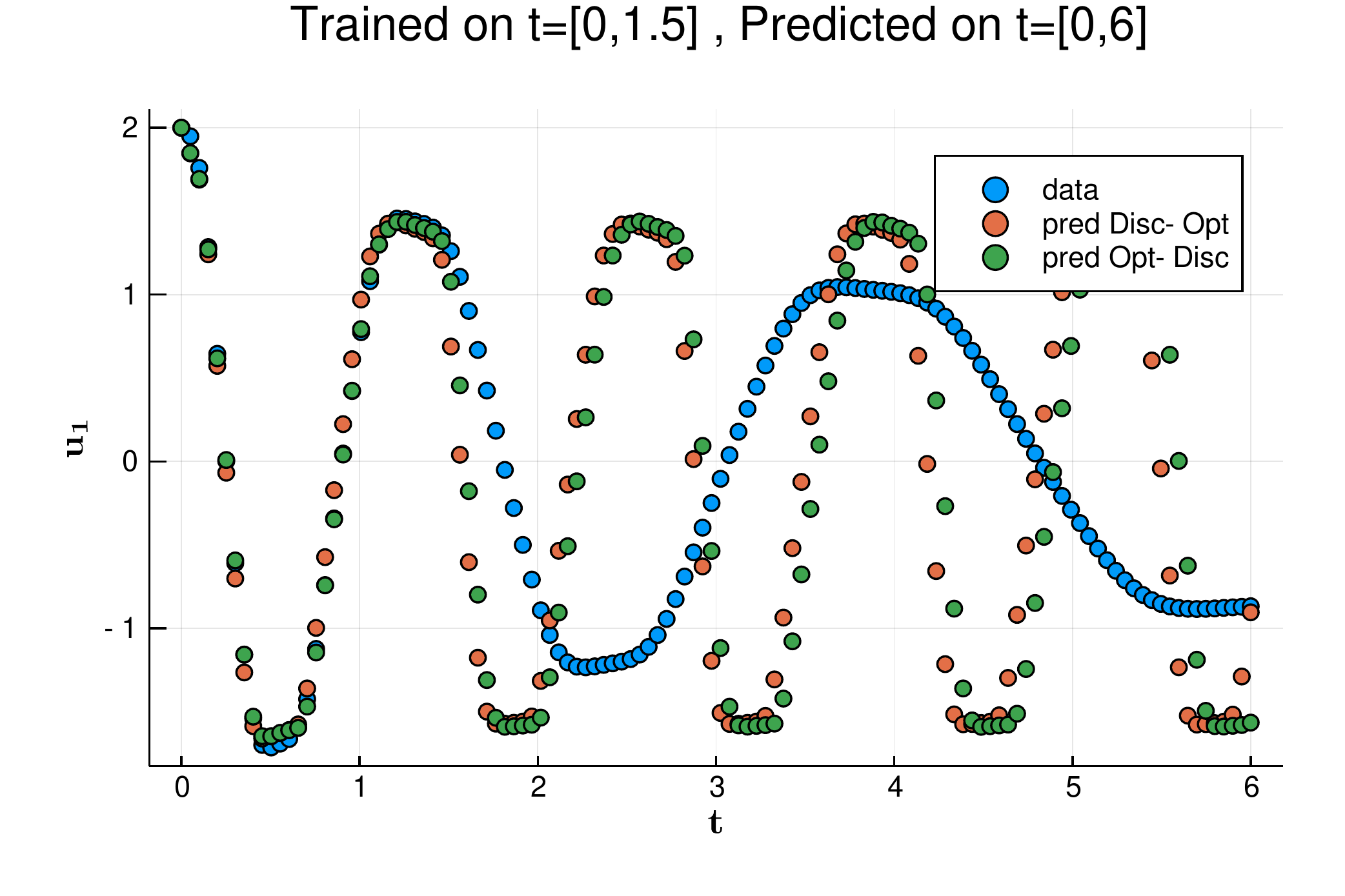}}\\
	\subfloat[Different seed 2]{\label{fig:seed400}%
	    	\includegraphics[width=0.3\linewidth]{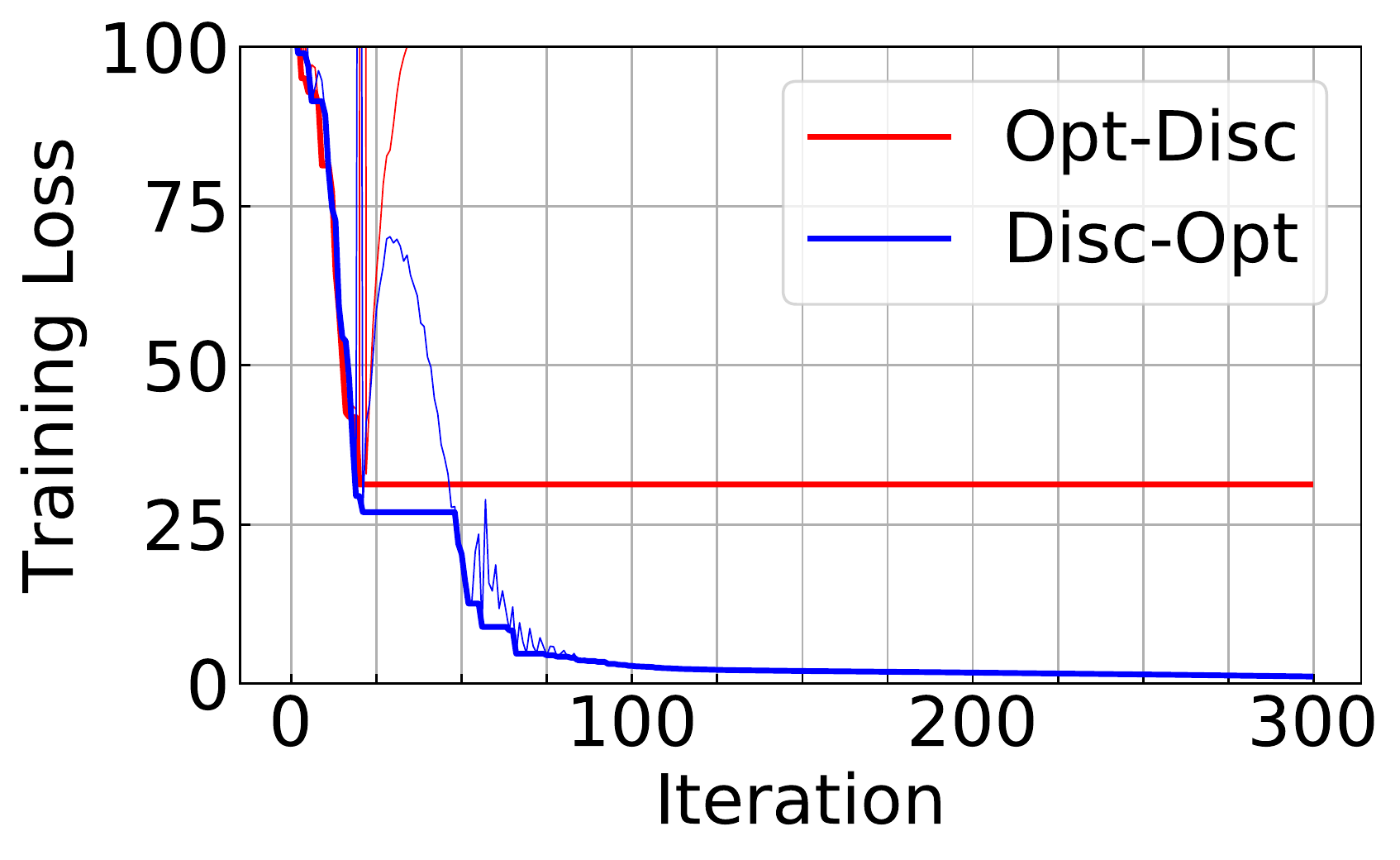}
	    	\includegraphics[width=0.3\linewidth]{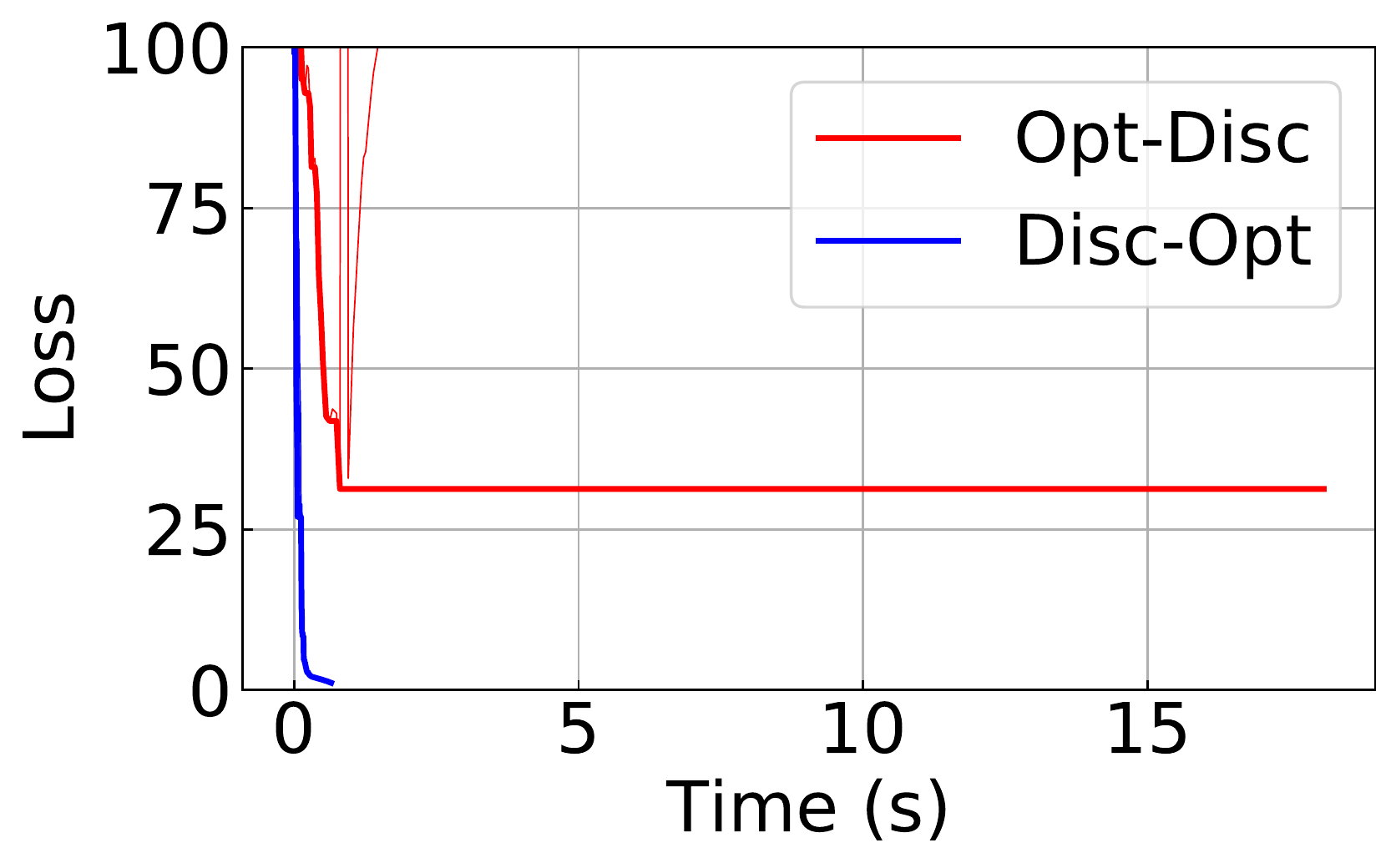}
			\includegraphics[width=0.3\linewidth]{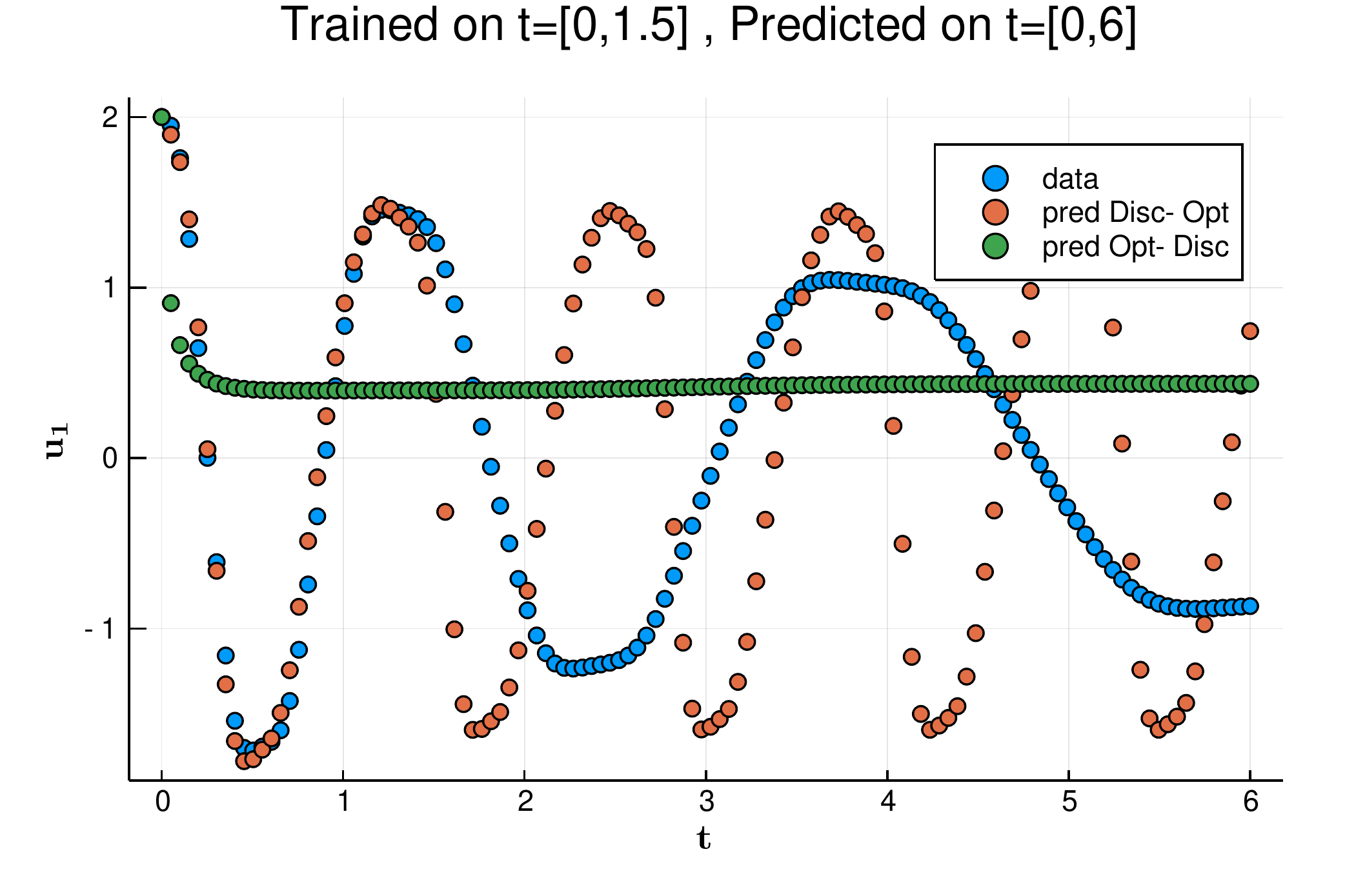}}\\
	\subfloat[Different seed 3]{\label{fig:seed0}%
	\includegraphics[width=0.3\linewidth]{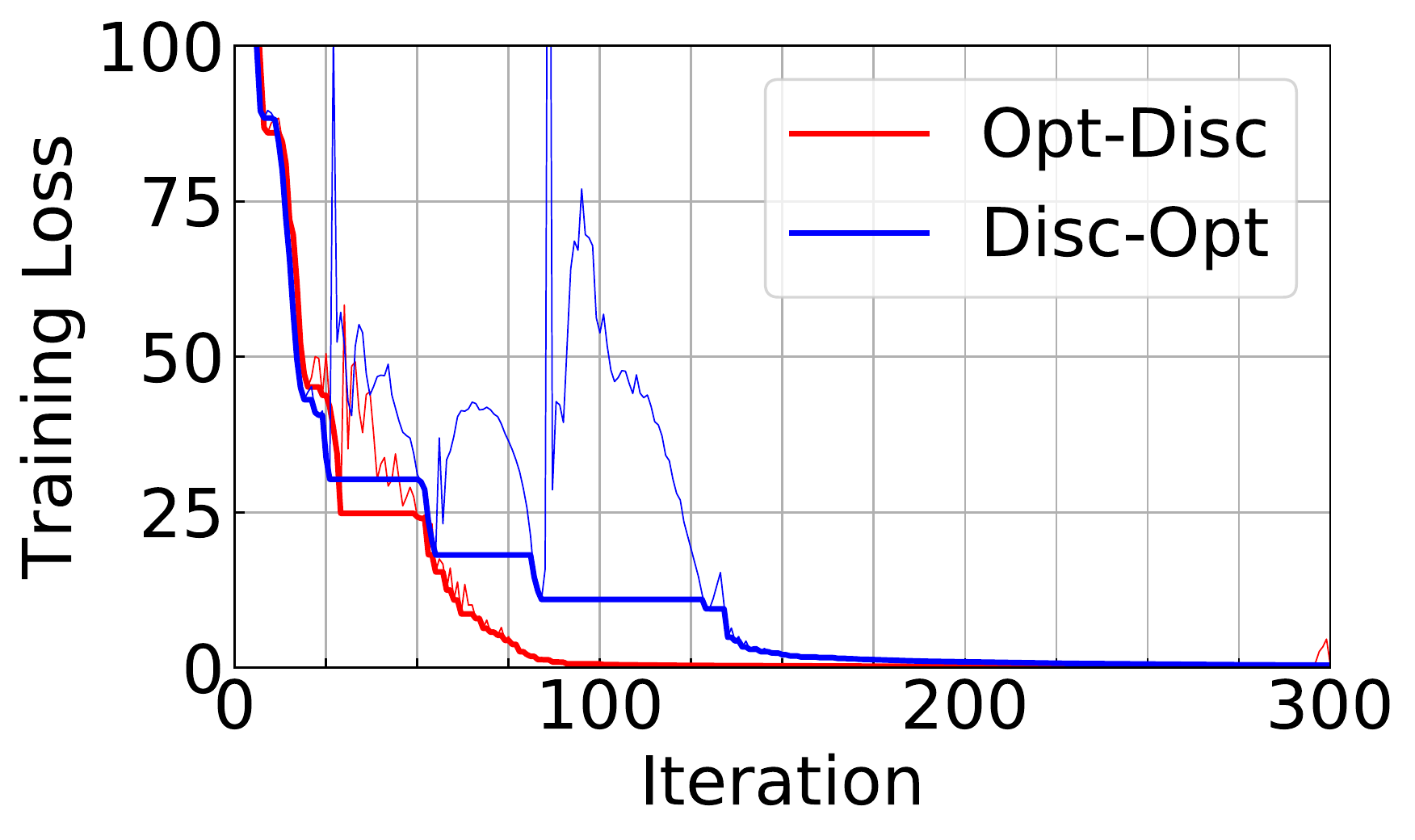}
	\includegraphics[width=0.3\linewidth]{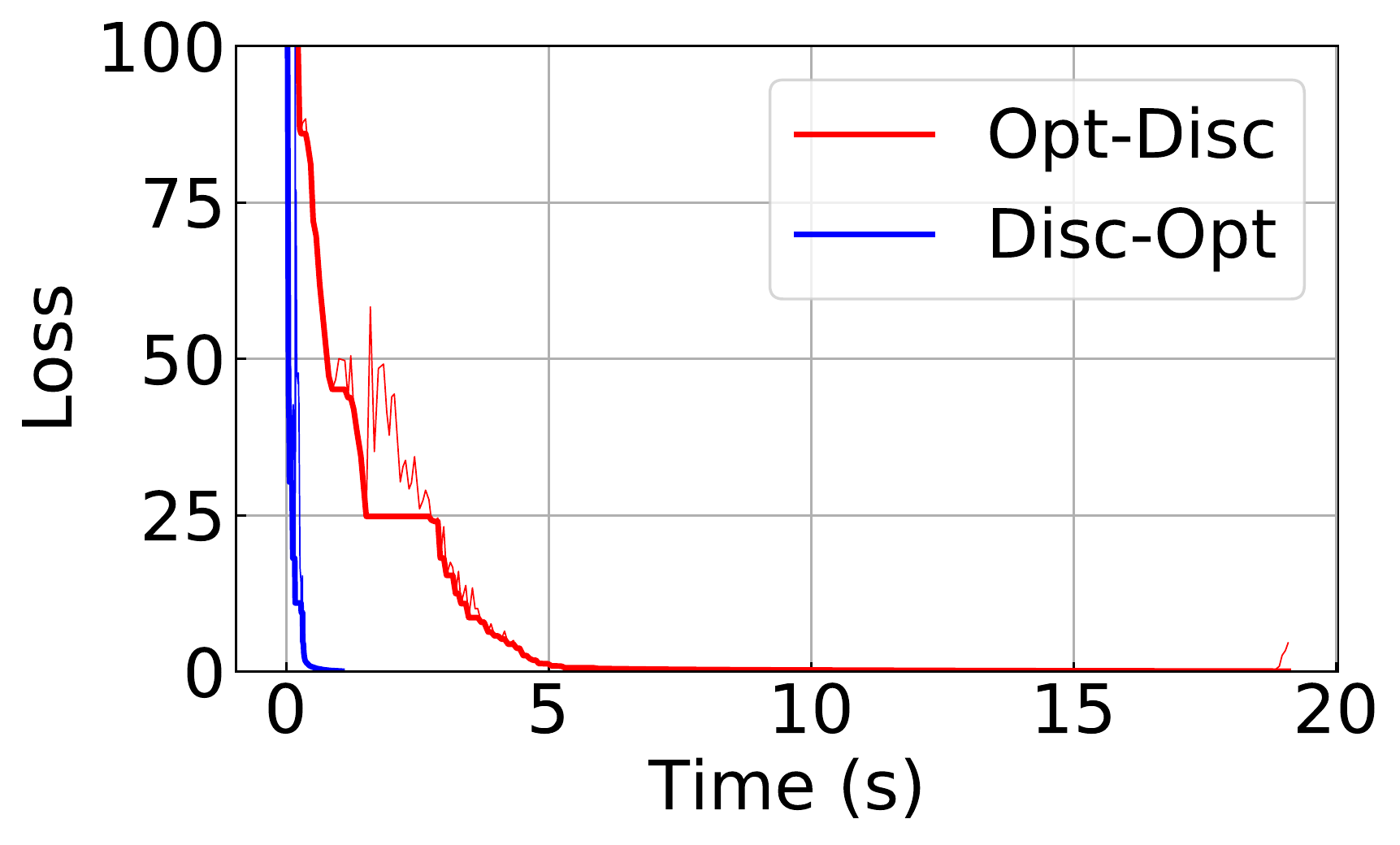}
			\includegraphics[width=0.3\linewidth]{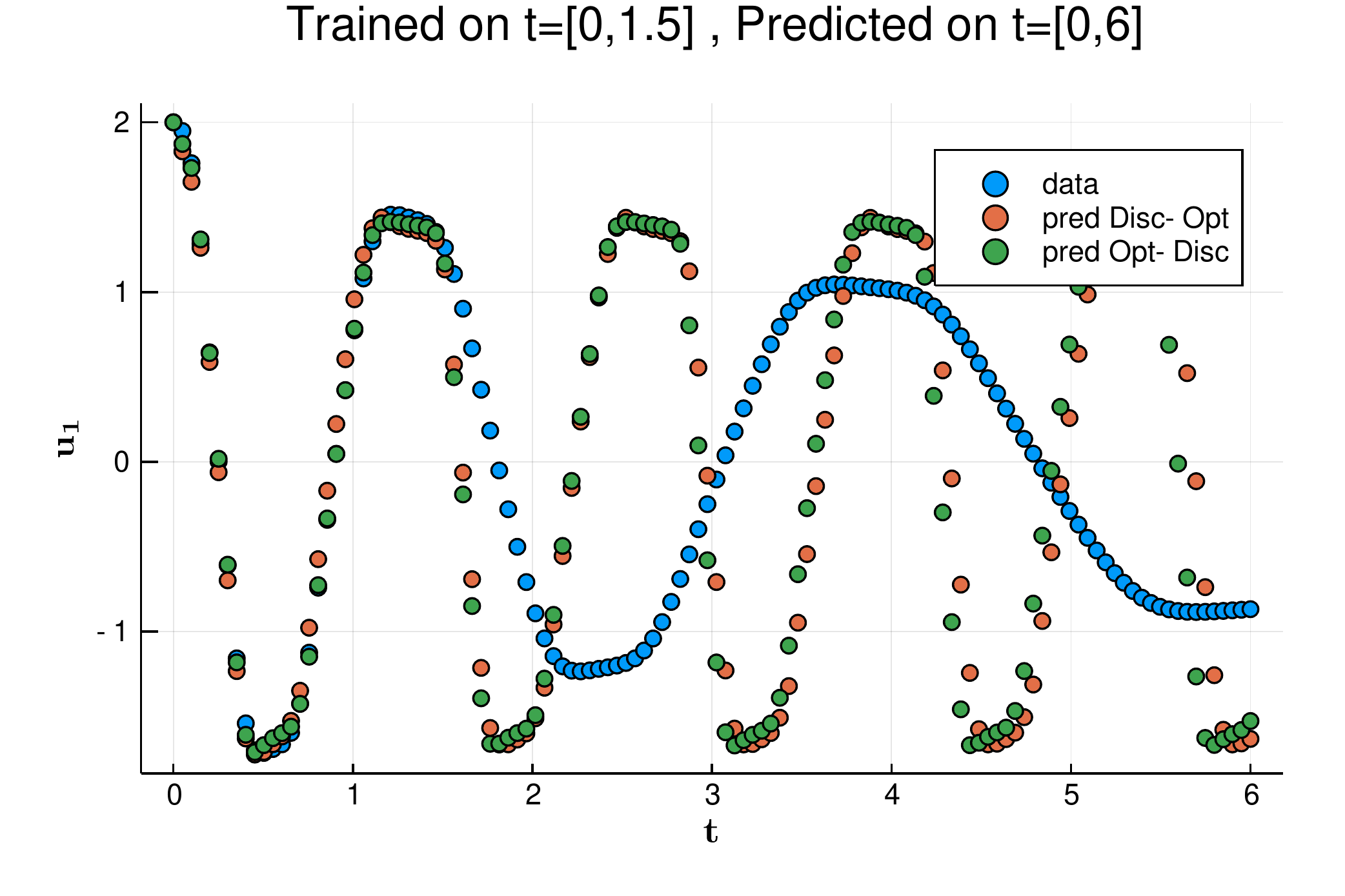}}\\
	
	}
	\caption{Time-series regression convergence for different initial parameterizations (cf. Figure~\ref{fig:timeAndLoss},~\ref{fig:extrapolation})}
	\label{fig:diff_seeds}
	\end{figure*}

	We repeat the time-series regression for several initial conditions to demonstrate that the results span a broad set of convergent behaviors.
	We plot three other random seeds (Figure~\ref{fig:diff_seeds}).
	Sometimes the \dto{} and \otd{} approaches converge at a similar rate per iteration and can have similar extrapolations (Figure~\ref{fig:diff_seeds}a). Some cases exist where the loss skyrockets, and the models fail to fully converge in the 300 iterations (Figure~\ref{fig:diff_seeds}b). Still other cases appear where \otd{} converges to a lower loss than \dto{} after 300 iterations, but extrapolation shows that the two models learn a similar ODE (Figure~\ref{fig:diff_seeds}c). In such cases, we observe that the \dto{} model still trains more efficiently in terms of time. In fact, out of ten random initial conditions selected, the \dto{} approach used fewer iterations to converge than the \otd{} approach in nine of the comparisons. In the one model (Figure~\ref{fig:diff_seeds}c) where the \otd{} converges in fewer iterations than \dto{}, the \dto{} model converges to the same loss in 466 iterations and still reduced training time by 94\%. Training costs reductions due to the \dto{} model across these ten initial conditions ranged from 94\% to 99\% with an average reduction of 97\% (a 20x speedup).

\subsection{Density Estimation Using Continuous Normalizing Flows} \label{sec:cnf}
	
	Density estimation aims at constructing the probability density function from a finite number of randomly drawn samples. 
	Many approaches exist for this problem including probabilistic models, variational autoencoders~\citep{kingma2013auto}, and autoregressive flows~\citep{gregor2014deep}.
	Here, we consider the {\em normalizing flows}~\citep{rezende2015} approach, which can be formulated using neural ODEs. 
	 
	Normalizing flows parameterize a path for the samples from the unknown density
	to the target density.
	The latter is usually chosen to be a standard normal density, hence the name normalizing flow.
	In continuous normalizing flows (CNFs), the model is a dynamical system formulated by a neural ODE in artificial time~\citep{chen2018neural}.
	A key motivation for using a continuous flow model is to obtain a generative model when the network is invertible.
	Computing the flow backward in time, one can compute the density of and compute samples from the unknown density. Furthermore, the continuous view allows for free-form network architectures~\citep{grathwohl2018ffjord}.
	 
	The CNF is a smooth bijective mapping between unknown density $\rho_0$ and known density $\rho_1$, is modeled by a neural ODE~\eqref{eq:f_euler}, and is trained using samples. In the discrete setting, normalizing flows map samples $\bfy_0 \sim \rho_0(\bfy_0)$ to points $\bfy_N$ that follow $\rho_1$.    
	In the continuous framework~\eqref{eq:cont_ode}, the CNF maps $\bfy(0)$ to $\bfy(T) = f(\bfth,\bfy(0))$.
	For implementation ease, we forward propagate $f$ from unknown density $\rho_0$ to the standard normal density $\rho_1$ (Figure~\ref{fig:cnf}).

	Consider the likelihood $p(\bfy(t))$ where $p \colon \mathbb{R}^{n_f} \rightarrow \mathbb{R}$. The likelihood $p(\bfy(t))$ varies continuously through time with features $\bfy$ following the neural ODE~\eqref{eq:cont_ode}. 
	The initial likelihood $p(\bfy(0))$ is the unknown density $\rho_0(\bfy(0))$, and we want $p(\bfy(T))=p(f({\bfth},\bfy(0)))$ to approximate $\rho_1(\bfy(T))$.  
	From this initial condition and the instantaneous change of variables~\citep{chen2018neural}, we obtain the initial value problem
  	\begin{equation} \label{eq:cnf_ivp}
  		\begin{aligned}
  			\partial_t [ \, \log p(\bfy(t)) - \log \rho_0(\bfy(0)) \, ] &= - \textrm{Tr} \, \, \nabla_{\bfy(t)} \ell\big(\bfth(t),\bfy(t),t\big) \\
  			p(\bfy(0)) - \rho_0(\bfy(0)) &= 0 .
  		\end{aligned}
  	\end{equation}

	To calculate how the likelihood varies over the time domain $[0,T]$, we integrate the log-likelihood ODE~\eqref{eq:cnf_ivp} through time and rearrange to
	\begin{equation*} 
	 	\log p(f(\bfth,\bfy(0))) = \log \rho_0(\bfy(0)) - \int_0^T \textrm{Tr} \, \, \nabla_{\bfy(t)} \ell\big(\bfth(t),\bfy(t),t\big) \, \du t .
	\end{equation*}
	We want $\log p(f(\bfth,\bfy(0)))$ to match $\log \rho_1$ where $\rho_1$ is the standard normal density. 
	Hence, the loss function, motivated by the Kullback–Leibler divergence~\citep{rezende2015}, is given by the negative log-likelihood of the final state as
	\begin{equation} \label{eq:cnf_loss}
		\calL(\bfth,\bfy)
		= \frac{n_f}{2} \log (2 \pi) + \frac12 \|\bfy(T)\|^2 - \int_0^T \textrm{Tr} \, \, \nabla_{\bfy(t)} \ell\big(\bfth(t),\bfy(t),t\big) \, \du t .
	\end{equation}

	We augment our initial ODE formulation~\eqref{eq:cont_ode} with the time integration~\eqref{eq:cnf_ivp}. Specifically, we vertically concatenate the states $\bfy$ with the log likelihood $p(\bfy(0))$ and solve both as a single initial value problem with the ODE solver~\citep{grathwohl2018ffjord}. As before, we minimize the loss $\calL(\bfth,\bfy)$ with respect to $\bfth$.
	A meaningful model must be invertible, i.e., the reverse mode must also push-forward $\rho_1$ to $\rho_0$, which is not enforced in training. We stress the importance of checking model invertibility (Tables~\ref{tab:8gaussians},~\ref{tab:flows}).

		\begin{table}
		\begin{center}
			\begin{tabular}{lllcccc} 
			\toprule
			\multirow{2}{*}{Data Set} & \multirow{2}{*}{$T$} & \multirow{2}{*}{Model}	& 
					Testing & \multirow{2}{*}{$\frac{\text{Time}}{\text{Epoch}}$ (s)} 	& Average 	& Inverse \\
		 & 	& & Loss 	& 											& NFE-F 	& Error \\
			\midrule
			\multirow{2}{*}{\textbf{Gaussian}} & \multirow{3}{*}{0.5} &
				Opt-Disc (dopri5) 		& 2.83 & 0.79 & 65 & 8.88e-7 \\
			\multirow{2}{*}{\textbf{Mixture}}	& & Disc-Opt (rk4 $h=0.05$) & 2.79 & 0.48 & 40 & 1.82e-8  \\
				& & Disc-Opt (rk4 $h=0.25$) & 2.83 & 0.10 & 8 & 1.47e-2  \\
			\bottomrule
			\end{tabular}	
		\end{center}
		
		\caption{CNF results for the Gaussian Mixture problem. The \dto{} approach with  $h=0.25$ has the lowest training time and achieves a spuriously low loss value; however, because the step size is too large, the discrete model loses its invertibility (Section~\ref{sec:too_coarse}). Decreasing the time step to $h=0.05$ still leads to a substantial reduction in the number of function evaluations, comparable loss, and comparable inverse error than the \otd{} approach.}
		\label{tab:8gaussians}
		\end{table}%

	\subsubsection{Normalizing Flow for Gaussian Mixture} \label{sec:sub_8gauss}
	To help visualize CNFs, consider the synthetic test problem where $\rho_0$ is the Gaussian mixture obtained by averaging eight bivariate Gaussians situated in a circular pattern about the origin (Figure~\ref{fig:cnf}). 
	For the \dto{} example, we use a Runge-Kutta 4 model with constant step size of $h{=}1$ and $T{=}2$. 
	We discretize the control layers at $\bfth(0)$ and $\bfth(1)$, while the state layers are discretized at one-quarter intervals as determined by the Runge-Kutta 4 scheme. 
	Overall, the forward pass requires eight evaluations of $\ell$ and the weights at intermediate time points are obtained by interpolating the two control layers (illustration in Figure~\ref{fig:cnf}).
	
	We use the neural ODE proposed in~\citet{grathwohl2018ffjord}, where $\ell$ is given by 
	\begin{equation*}
		\ell(\bfth(t),\bfy(t),t) = c_{64, 2}(\bfth_4(t),\cdot,t) \circ c_{64, 64}(\bfth_3(t),\cdot,t) \circ c_{64, 64}(\bfth_2(t),\cdot,t) \circ c_{2, 64}(\bfth_1(t),\bfy(t),t). 
	\end{equation*}
	The so-called  concatsquash layer $c_{i, j}(\bfth(t),\bfy(t),t)$ maps features $\bfy(t) \in \mathbb{R}^i$ to outputs in $\mathbb{R}^j$ and is defined as
	\begin{equation} \label{eq:concatsquash}
		c_{i, j}(\bfth,\bfy,t) = ( \myD{2} \bfy) (\sigma \circ \myD{1} t  ) + (\myD{0} t ).
	\end{equation}
	Here, the nonlinear activation function $\sigma$ is the hyperbolic tangent and $\myD{0}$ uses no bias. This layer accepts and operates on the space features and the time. 

	To solve the problem, the CNF needs to be trained to push the Gaussian mixture $\rho_0$ to the center to form $\rho_1$ (Figure~\ref{fig:cnf}).
	To discretize the loss function, we use samples drawn from the Gaussian mixture $\rho_0$. 
	The reverse mode of the model will separate the large Gaussian back into the eight terms of the Gaussian mixture. 
	What makes the problem challenging is that the true inverse is discontinuous at the origin.
	
	To gauge the accuracy of the trained flow model, we compute the inverse error
	\begin{equation} \label{eq:inverse}
		\Delta_{\rm inv}(\bfy_0) = \lVert f^{-1} \left( \bfth , f(\bfth, \bfy_0) \right) - \bfy_0 \rVert, 
	\end{equation}
	where $\bfy_0$ is  sampled from $\rho_0$. This calculates the Euclidean distance between an initial point $\bfy_0$ and the result from mapping $\bfy_0$ forward then back.
	 We report the average Euclidean distance for the samples in one batch (Tables~\ref{tab:8gaussians},~\ref{tab:flows}); the \dto{} and \otd{} approaches use the same batch.
	 
	A relatively small network model consisting of one control layer accurately solves the problem (Table~\ref{tab:8gaussians}).
	For a fixed 10,000 iterations, the \dto{} approach with $h=0.05$ demonstrates a slight speedup per iteration and achieves a similar testing loss as the \otd{} approach; also, the \dto{} solution has a comparable inverse error.

\subsubsection{Too Coarse Time Stepping} \label{sec:too_coarse}

	Using a too large step size, $h=0.25$, the \dto{} model may achieve a low testing loss while losing the invertibility.
	In fact, this flow fails to accurately align the densities.

	A sufficiently small step size is needed to ensure the discrete forward propagation adequately represents the neural ODE~\citep{ascher2011textbook}; however, a too small step size leads to excessive and unnecessary computation and training time. Hence, there is a trade-off between accuracy and speed.  For the Gaussian mixture problem, we trained two \dto{} networks with step sizes $h=0.05$ and $h=0.25$ (Table~\ref{tab:8gaussians}). The $h=0.25$ step size showed an inverse error much higher than the other two because the larger step size defines a coarser state grid, which leads to a less-accurate approximation for the ODE~\eqref{eq:cont_ode}.
	
\begin{figure*}
	  \centering
	  {%
	    \subfloat[\dto{} RK4 with $h=0.05$\label{fig:a}]{%
	      \includegraphics[clip,trim=1cm 1cm 1cm 0cm,width=0.7\linewidth]{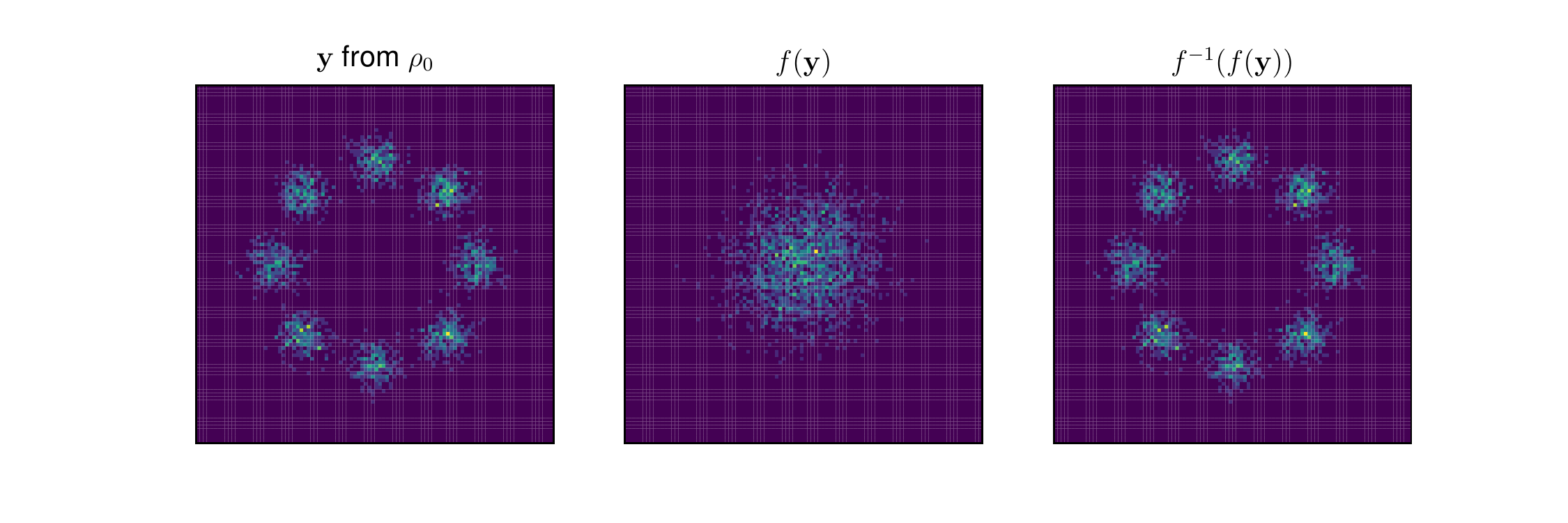}}\\
	    \subfloat[\dto{} RK4 with $h=0.25$\label{fig:b}]{%
	      \includegraphics[clip,trim=1cm 1cm 1cm 0cm,width=0.7\linewidth]{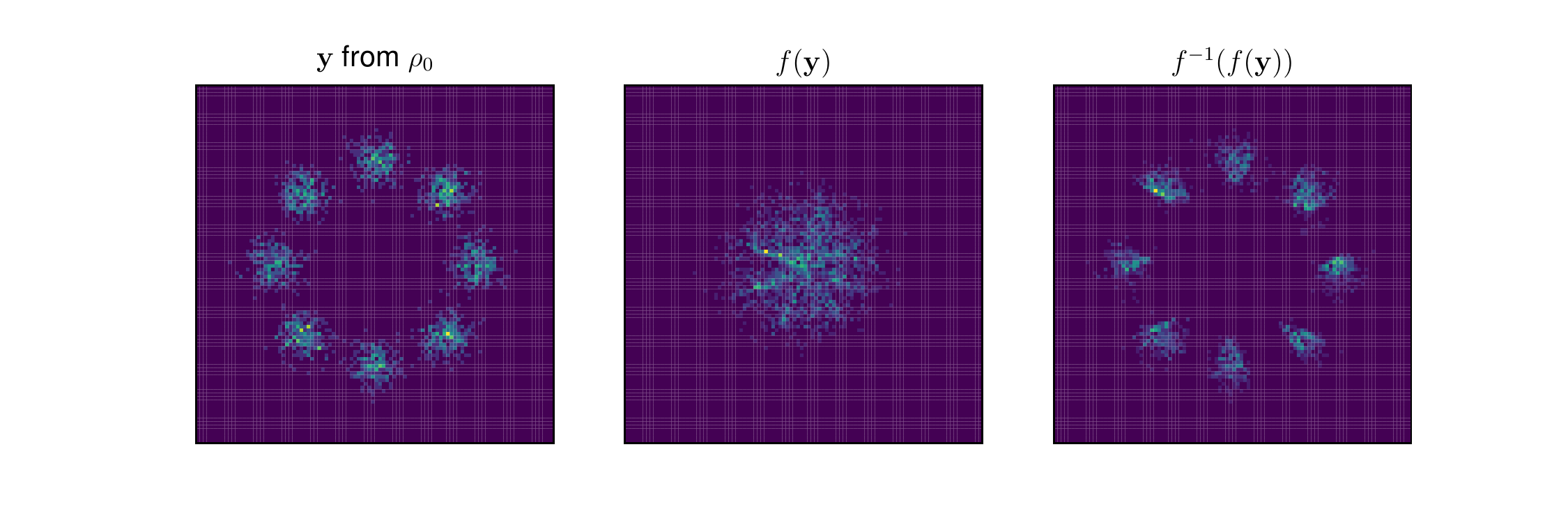}}\\
	    \subfloat[Testing for \dto{} RK4 with $h=0.05$ (but the model was trained on RK4 with $h=0.25$)]{\label{fig:c}%
	      \includegraphics[clip,trim=1cm 1cm 1cm 0cm,width=0.7\linewidth]{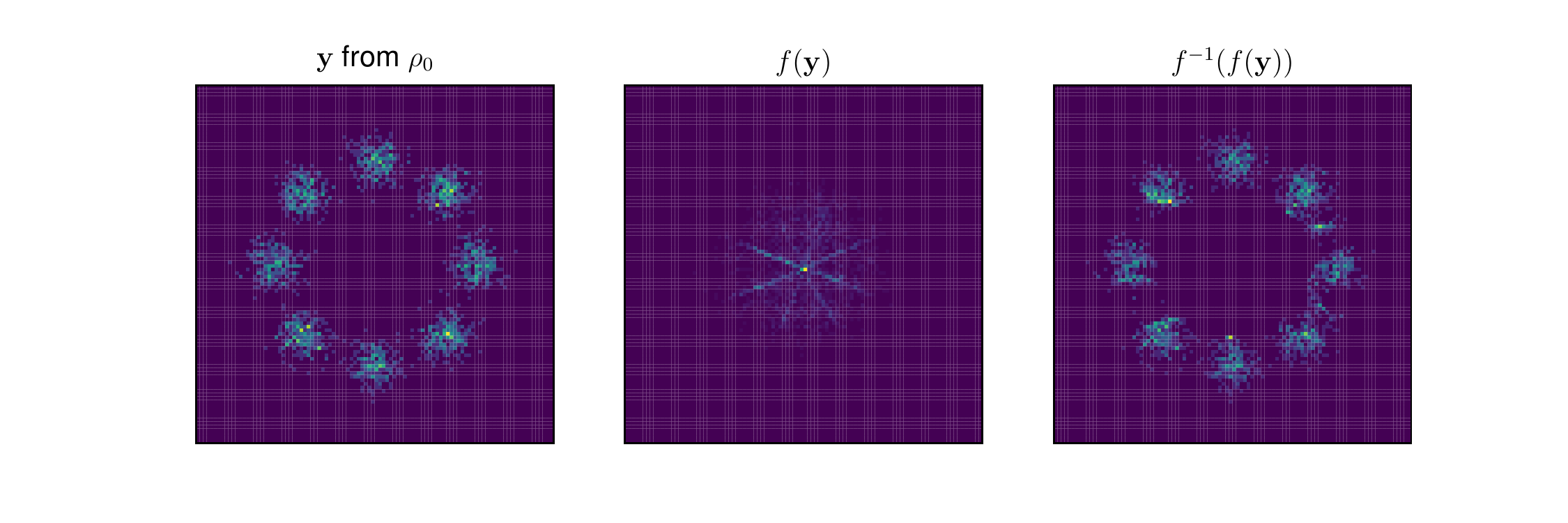}}\\
 	    \subfloat[Invertibility of \dto{}  RK4 with $h=0.25$ for one point $\hat{\bfy}(0)$]{\label{fig:d}%
 	      \includegraphics[width=0.7\linewidth]{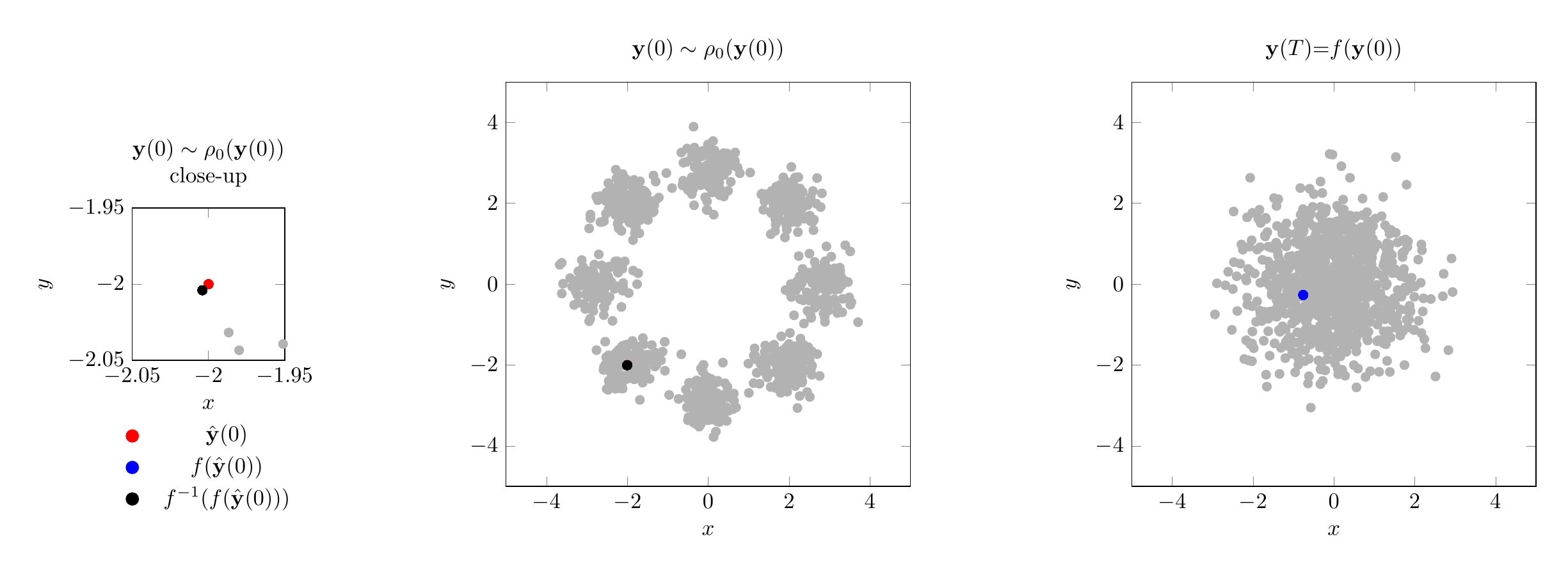}}
	  }
	  \caption{Using a step size not sufficiently small enough in the \dto{} method can result in lack of invertibility. (a) The fine \dto{} approach demonstrates a good model. (b) The coarse model fails to make a smooth Gaussian and has some inaccuracies with the inverse. (c) We can switch the testing solver for that coarse model to mitigate some of the issues. (d) A representation of the point $\bfy(0) = (-2,-2)$ through the propagation shows the scale of the inverse error. The inverse error of this point is less than the average for the model (Table~\ref{tab:8gaussians}) so we expect that other points demonstrate a greater inverse error.}
	  \label{fig:coarse8gaussian}
	\end{figure*}

	The coarse model ($h=0.25$) achieves a loss on par with the fine grid models and an inverse error that is accurate for a couple decimal points (Table~\ref{tab:8gaussians}). However, visually, the forward propagation leaves artifacts and does not create a smooth Gaussian (Figure~\ref{fig:coarse8gaussian}b).
	Also, the inverse flow does not fully match the initial distribution. Therefore, we observe issues with a coarse discrete model that are not indicated by the loss.
	Re-discretizing the coarse model after training, we can switch to a finer solver for testing (Figure~\ref{fig:coarse8gaussian}c), which smooths some of the artifacts in the central Gaussian and curves the inverse plot. The coarse model without changes (Figure~\ref{fig:coarse8gaussian}b) may be sufficient for some application because the inverse error could be sufficiently small (Figure~\ref{fig:coarse8gaussian}d).

	\subsubsection{High-Dimensional CNF Examples}
	We now consider some common higher-dimensional instances of the CNF problem arising in density estimation; the public data sets we use are listed in Table~\ref{tab:flows}.
	A detailed description of the data sets is given in \citet{papamakarios2017masked}. 
	In short, \textsc{Miniboone}, \textsc{Power}, \textsc{Hepmass}, and \textsc{Gas} are data sets compiled for various tasks and housed by the University of California, Irvine (UCI) Machine Learning Repository. The contexts include neutrinos, electric power consumption of houses, particle-producing collisions, and gas sensors, respectively. 
	The Berkeley Segmentation Data Set (\textsc{Bsds300}) contains image patches for image segmentation and boundary detection. \citet{papamakarios2017masked} preprocessed these data sets for use in density estimation. These data sets have also been used to validate the \otd{} method in~\citet{grathwohl2018ffjord}.
	
	Given the samples in the data sets, the goal in CNFs is to estimate their underlying density $\rho_0$ by constructing a mapping to the standard normal density with density $\rho_1$.
	We quantify this using the loss~\eqref{eq:cnf_loss} computed on a hold-out testing set. 
	Since the goal is to invert the CNF to characterize $\rho_0$, we numerically evaluate the inverse error (Table~\ref{tab:flows}).
	
	For each data set, we define an $\ell$ via concatenations of concatsquash layer $c_{i , j}(\bfth(t),\bfy(t),t)$~\eqref{eq:concatsquash} that are more complicated than for the Gaussian mixture problem. For hyperparameters, we can select an activation function $\sigma$, number of hidden layers $n_h$, number of flow steps $s$, and the number of hidden dimensions, which we choose based on some multiplier $m$ of the number of input features (Table~\ref{tab:hyperparameters}). Our general layer $\ell$ then is 
	\begin{equation*}
		\ell(\bfth(t),\bfy(t),t) = \left[ c_{mn_f , n_f}(\bfth(t),\cdot,t) \circ c_{mn_f , mn_f}(\bfth(t),\cdot,t)^{n_h{-}1} \circ c_{n_f , mn_f}(\bfth(t),\bfy(t),t) \right]^{s}.
	\end{equation*}
	Each of the $s$ blocks starts with one concatsquash layer that maps from $\mathbb{R}^{n_f}$ to $\mathbb{R}^{mn_f}$, then $n_h{-}1$ more concatsquash layers that maintain the dimensionality, then a final concatsquash layer that maps back to $\mathbb{R}^{n_f}$. Depending on the difficulty of the data set, different hyperparameters are selected; we mostly followed the hyperparameters (Table~\ref{tab:hyperparameters}) tuned by FFJORD~\citep{grathwohl2018ffjord}.

	The large data sets have higher dimensionality. As such, computing the full trace~\eqref{eq:cnf_ivp} during training becomes onerous. Using Hutchinson's trace estimator~\citep{hutchinson1989stochastic}
	\begin{equation*}
		\textrm{Tr}(A) = \mathbb{E}_{\phi(\epsilon)} \left[ \epsilon^{\top}A \epsilon\right] 
	\end{equation*}
	for square matrix $A$ and noise vector $\epsilon$ with density $\phi(\epsilon)$ where $\mathbb{E}[\epsilon]=0$ and $\textrm{Cov}(\epsilon)=I$, we can make training less computationally expensive and less time-consuming~\citep{grathwohl2018ffjord}. During testing, the model computes the full trace without estimation. The \otd{} approach uses trace estimation and a noise vector $\epsilon$ resampled for each batch. For fair comparison, we adhere to these design decisions for \dto{}.

	The performance on the testing sets of all data sets demonstrates that the \otd{} and \dto{} approaches result in similar negative log-likelihood loss (Table~\ref{tab:flows}). However, \dto{} does so more quickly on most of the data sets but with greater inverse error. Although the automatic differentiation used in \dto{} is faster than the adjoint-based recalculation used in \otd{} for batches of the same size, the memory requirements of storing intermediates in automatic differentiation restrict the \dto{} batch sizes to be much smaller than the \otd{} batch sizes. Checkpointing can also handle the memory constraints. Nonetheless, the \otd{} method for solving the \textsc{Hepmass} density estimation trains faster than \dto{} because of the larger batches. Since we use the same models, code, and hyperparameters (Table~\ref{tab:hyperparameters}) as FFJORD~\citep{grathwohl2018ffjord}, we witness similar testing losses (Table~\ref{tab:CNF_other_methods}).

	We see \dto{} converge in less time for most data sets, but no timing payoff for \textsc{Hepmass} (Table~\ref{tab:flows}). Even so, on average for the five high-dimensional data sets, \dto{} offers a 6x speedup in training over the \otd{} method.
	All \otd{} and \dto{} models are competitive in performance with other density estimation models (Table~\ref{tab:CNF_other_methods}).

 	\begin{table}[t]
 		\centering
 		\begin{tabular}{lllccccc} 
 		\toprule
 		\multirow{2}{*}{Model} & Training & Inference & Total Training  & Training & Inference & Testing & Inverse \\
 		& Solver & Solver & Time (s) & Avg NFE-F & NFE-F & Loss & Error \\
 		\midrule
 		\multicolumn{8}{c}{\textbf{POWER}, $T{=}5$} \\ 
 			Opt-Disc & dopri5 	     & dopri5	& 272K 	& 511 & 649  & -0.42  & 1.98e-7\\
   			Disc-Opt & rk4 $h{=}0.10$ & dopri5 	& 56.0K	& 200 	& 2066 & -0.25 	& 1.58e-5\\
 			Disc-Opt & rk4 $h{=}0.10$ & rk4 $h{=}0.10$  	& 56.0K 	& 200   & 200  & -0.33  & 4.23e-3  \\
 		\midrule
 		\multicolumn{8}{c}{\textbf{GAS}, $T{=}5$} \\ 
 			Opt-Disc & dopri5 	   		& dopri5		& 205K & 454 & 527 & -10.53 & 2.71e-7 \\
   			Disc-Opt & rk4 $h{=}0.10$ 	& dopri5		& 121K & 200   & 437 & -10.27 & 4.71e-7\\
 			Disc-Opt & rk4 $h{=}0.10$ 	& rk4 $h{=}0.10$ & 121K & 200   & 200 & -10.27 & 6.63e-3\\
 		\midrule
 		\multicolumn{8}{c}{\textbf{HEPMASS}, $T{=}2$} \\
 			Opt-Disc & dopri5 	   		& dopri5		  & 186K & 739 & 866 	& 16.61 & 9.26e-7\\
 			Disc-Opt & rk4 $h{=}0.10$ 	& dopri5		  & 281K	& 400	& 765   & 16.60 & 1.09e-6\\
  			Disc-Opt & rk4 $h{=}0.10$ 	& rk4 $h{=}0.10$  & 281K & 400	& 400 	& 16.60 & 2.25e-4\\
 		\midrule
 		\multicolumn{8}{c}{\textbf{MINIBOONE}, $T{=}1$} \\
 			Opt-Disc & dopri5 	   		& dopri5		& 36.1K & 113 & 132 & 10.64 & 1.85e-7\\
 			Disc-Opt & rk4 $h{=}0.25$ 	& dopri5		& 3.77K & 16    & 118 & 10.50 & 2.17e-7\\
 			Disc-Opt & rk4 $h{=}0.25$ 	& rk4 $h{=}0.25$& 3.77K & 16 	  & 16 	& 10.45	& 1.45e-3 \\
 		\midrule
 		\multicolumn{8}{c}{\textbf{BSDS300}, $T{=}2$} \\
 			Opt-Disc & dopri5 	   		& dopri5		& 767K & 377 & 544 	& -142.91 & 1.06e-7\\
 			Disc-Opt & rk4 $h{=}0.05$ 	& dopri5		& 49.7K & 160   & 511 	& -146.23 & 1.82e-7 \\
   			Disc-Opt & rk4 $h{=}0.05$ 	& rk4 $h{=}0.05$& 49.7K & 160	  & 160 	& -146.14 & 3.75e-4\\
 		\bottomrule
 		\end{tabular}	
 	\caption{CNF density estimation on large public data sets. Negative log-likelihood loss (lower is better) in nats (natural unit of information). The number of functions evaluations for the forward propagation (NFE-F) are averaged over all the batches and epochs. 
 	Performance is competitive with other methods (Table~\ref{tab:CNF_other_methods}). The \otd{} approach for \textsc{Bsds300} was terminated prematurely after nearly nine days of training. 
	While the discrete model in \dto{} can have a large inverse error with the step size used in training,
	re-discretizing in the inference phase can reduce this error substantially. In summary, \dto{} reduces the training times and yields invertible models with comparable performance to \otd{}. } 
 	\label{tab:flows}
 	\end{table}

	\begin{table}
	\begin{center}
		\begin{tabular}{lccccccc} 
		\toprule			
		\multirow{2}{*}{Data Set} & \multirow{2}{*}{Model} & \multirow{2}{*}{step size}	& \multirow{2}{*}{batchsize} & \multirow{2}{*}{nonlinearity} &
			 \multirow{2}{*}{\# layers}  & hidden dim 	& \# flow\\
	 		 &   &   &  & 	&			 & multiplier 	& 	steps	\\
		\midrule
		\multirow{2}{*}{\textbf{POWER}} 	& Opt-Disc	& -    & 30000 	& \multirow{2}{*}{tanh} 	& \multirow{2}{*}{3} & \multirow{2}{*}{10} 	& \multirow{2}{*}{5} \\
											& Disc-Opt	& 0.10 & 10000 	& & & & \\
		\midrule
		\multirow{2}{*}{\textbf{GAS}} 		& Opt-Disc	& -    & 5000 	& \multirow{2}{*}{tanh} 	& \multirow{2}{*}{3} & \multirow{2}{*}{20} & \multirow{2}{*}{5} \\
											& Disc-Opt	& 0.10 & 5000 	& & & & \\	
		\midrule
		\multirow{2}{*}{\textbf{HEPMASS}} 	& Opt-Disc	& -    & 10000 	& \multirow{2}{*}{softplus}	& \multirow{2}{*}{2} & \multirow{2}{*}{10} 	& \multirow{2}{*}{10}\\
											& Disc-Opt	& 0.10 & 5000 	& & & & \\
		\midrule
		\multirow{2}{*}{\textbf{MINIBOONE}} & Opt-Disc	& -    & 5000 	& \multirow{2}{*}{softplus} & \multirow{2}{*}{2} & \multirow{2}{*}{20} 	& \multirow{2}{*}{1} \\
											& Disc-Opt	& 0.25 & 5000 	& & & & \\
		\midrule
		\multirow{2}{*}{\textbf{BSDS300}} 	& Opt-Disc	& -    & 10000 	& \multirow{2}{*}{softplus} & \multirow{2}{*}{3} & \multirow{2}{*}{20} & \multirow{2}{*}{2} \\
											& Disc-Opt	& 0.05 & 500 	& & & & \\
		\bottomrule
		\end{tabular}	
	\end{center}
	\caption{ CNF hyperparameters. Mostly a replication of \citet{grathwohl2018ffjord}'s table. All used initial learning rate=1e-3. \dto{} often used smaller batch sizes so that the batch would fit into the Titan X GPU (memory limit: 12GB).}
	\label{tab:hyperparameters}
	\end{table}

	\begin{table}
		\begin{center}
			\begin{tabular}{lccccc} 
			\toprule			
							& POWER 	& GAS    &  HEPMASS	& MINIBOONE & BSDS300 \\
			\midrule
			Real NVP		& -0.17 	& -8.33  &  18.71	& 13.55 & -153.28 \\
			Glow			& -0.17 	& -8.15  &  18.92	& 11.35 & -155.07 \\
			FFJORD			& -0.46 	& -8.59  &  14.92	& 10.43 & -157.40 \\
			our FFJORD (Opt-Disc)& -0.42 	& -10.53 &  16.61	& 10.64 & -142.91 \\
			our FFJORD (Disc-Opt)& -0.25 	& -10.27 &  16.60	& 10.50 & -146.23 \\
			\midrule
			MADE			& 3.08 		& -3.56  &  20.98	& 15.59 & -148.85 \\
			MAF				& -0.24 	& -10.08 &  17.79	& 11.75 & -155.69 \\
			TAN				& -0.48 	& -11.19 &  15.12	& 11.01 & -157.03 \\
			MAF-DDSF		& -0.62 	& -11.96 &  15.09	& 8.86  & -157.73 \\

			\bottomrule
			\end{tabular}	
		\end{center}
	
	\caption{CNF comparison with other methods. Replicated from~\citet{grathwohl2018ffjord} with the addition of the two models we trained. Our \otd{} models should be the exact same as FFJORD; minor differences attributed to the randomness of batching and stochastic subgradient optimization methods. }
	\label{tab:CNF_other_methods}
	\end{table}

	\subsubsection{Re-Discretizing the Trained Flows}
	
	Re-discretization in the inference phase presents a straightforward idea for reducing the inversion error of \dto{} trained models.
	
	Using the \textsc{Miniboone} data set, we consider training a CNF with the \dto{} approach with \rk{} and $h{=}0.25$. 
	Training takes less than 3,800 seconds, roughly one-tenth of the time required by the \otd{} (Table~\ref{tab:flows}). 
	After training (e.g., during inference or model deployment) we re-discretize the model using step size $h{=}0.05$ (Table~\ref{tab:multilayer}). 
	This maintains a similar testing loss of 10.50 while substantially reducing the inverse error.
	Demonstrating the flexible choice of solver, we also consider the adaptive dopri5 solver in the evaluation (which has 118 function evaluations).
	
	Since the \dto{} model was trained using a fixed step size, the generalization with respect to re-discretization is remarkable; note the low loss observed for all time integrators.
	As to be expected, the inverse error is reduced for more accurate time integrators.
	Less accurate time integrators can use their error to their advantage and result in lower loss values while sacrificing invertibility. Therefore, both loss and invertibility should be used to evaluate CNF performance.

	\begin{table}
	\begin{center}
		\begin{tabular}{lcccccc} 
		\toprule
			& \multicolumn{3}{c}{\textbf{Trained via Disc-Opt}} & \multicolumn{3}{c}{\textbf{Trained via Opt-Disc}} \\ 	
				& Testing  & \multirow{2}{*}{NFE-F} & Inverse & Testing  & \multirow{2}{*}{NFE-F} & Inverse \\
				& Loss  & & Error & Loss  & & Error \\
		\midrule
		\textbf{MINIBOONE, $T=1$} & & & & &\\

		dopri5 rtol 1e-1 atol 1e-3 	& 10.075 & 28 	& 4.25e-2	& 10.108 	& 29 & 3.16e-2 \\	
		dopri5 rtol 1e-2 atol 1e-4 	& 10.529 & 38 	& 3.55e-3	& 10.669 	& 38 & 1.61e-2 \\		
		dopri5 rtol 1e-6 atol 1e-8 	& 10.502 & 118  & 2.17e-7	&  \textbf{10.636}  &  \textbf{132} & \textbf{1.85e-7} \\
		rk4 $h$=0.05 				& 10.502 & 80	& 3.40e-7	& 10.636 	& 80 & 	5.22e-7 \\
		rk4 $h$=0.25 				&  \textbf{10.454}& \textbf{16}	& \textbf{1.45e-3} &  10.590 & 16	& 2.57e-3 \\
		rk4 $h$=0.50 				&  10.122 & 8	& 2.97e-2	& 10.239	& 	8	& 4.79e-2 \\
		rk4 $h$=1.0  				&  8.916  & 4	& 2.77e-1	& 9.082		& 	4	& 3.70e-1 \\
		\bottomrule
		\end{tabular}	
	\end{center}
	
	\caption{Re-discretization of the trained CNF  in the  inference phase. 
	Model trained using the solver associated with \textbf{bold} values. All other results in the column result from changing the solver settings (and thus the state discretization) for the forward propagation when running on the test data. The inverse error $\Delta_{\rm inv}$ averaged over one batch via~\eqref{eq:inverse}. Since the \dto{} problem is solved with a fixed step size and not using adaptive integration used in \otd, the generalization to other discretization is remarkable. }
	\label{tab:multilayer}
	\end{table}

	\subsubsection{Multilevel Training}
	
	Models trained with finer grids tend to have better convergence than models trained with coarser grids; however, these fine grid models consume much more training time from the large NFE. Taking a \textit{multilevel} approach, we train the initial epochs with a coarse grid (few state layers), then add state layers to make the grid finer for the final training epochs.
	
\begin{figure*} 
	  	\centering
  			\begin{tabular}{lccc}
  			\toprule
  			Model & Total Training Time (s) & Testing Loss & Inverse Accuracy \\
  			\midrule
  			Opt-Disc 				& 23.8K & 10.64 & 1.85e-7\\
  			Opt-Disc Multilevel 	& 20.2K & 10.43 & 1.22e-7\\
  			Disc-Opt 				& 2.73K & 10.45 & 1.45e-3\\
  			Disc-Opt Multilevel 	& 2.38K & 10.38 & 1.32e-3\\
  			\end{tabular}
  			\includegraphics[clip, trim=0.0cm 0.2cm 0.0cm 0.0cm, width=0.295\linewidth]{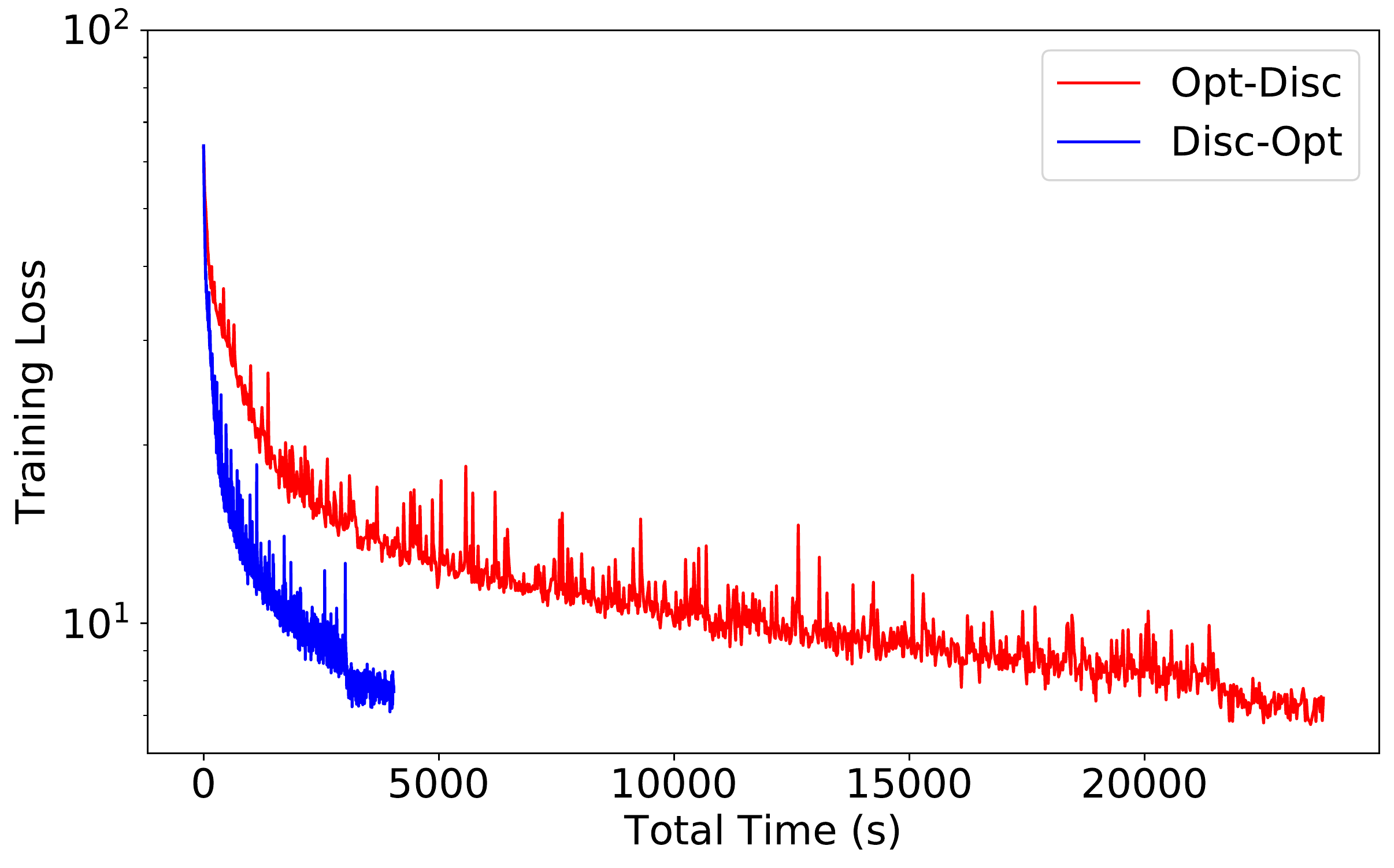}
			\includegraphics[clip, trim=1.1cm 0cm 1.8cm 1cm, width=0.3\linewidth]{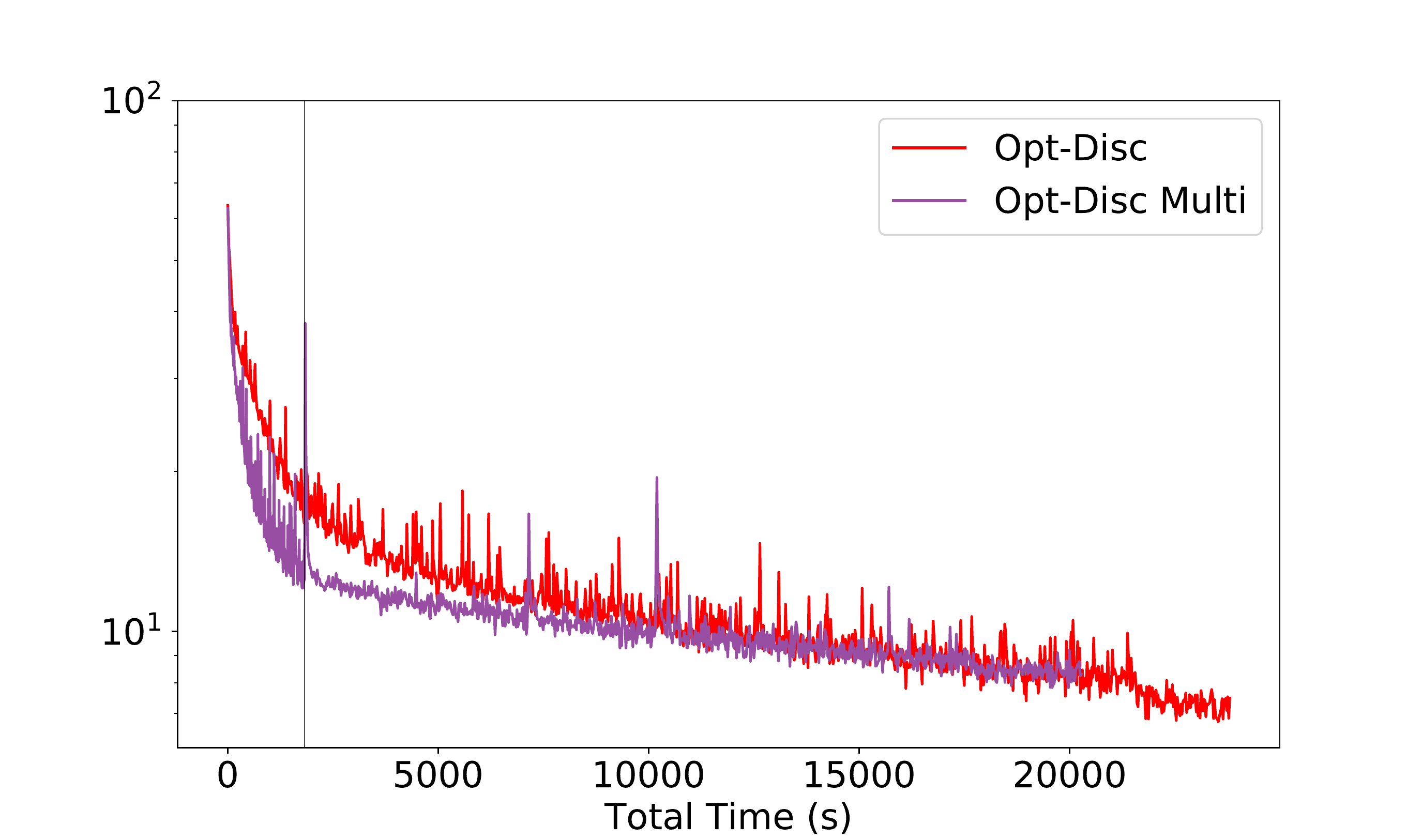}
			\includegraphics[clip, trim=1.1cm 0cm 1.8cm 1cm, width=0.3\linewidth]{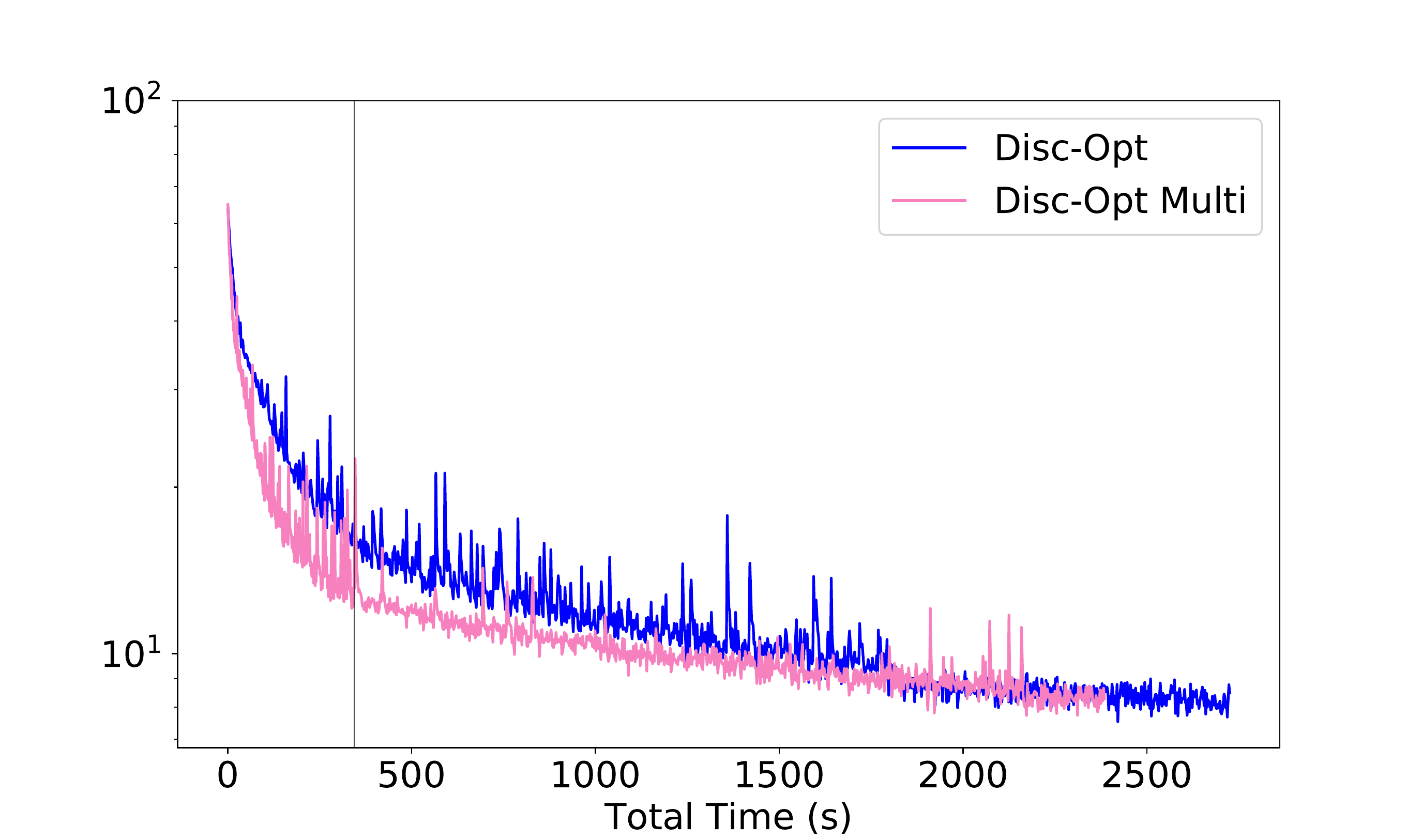}
			\caption{Training CNFs on \textsc{Miniboone} with and without multilevel. For multilevel, train first 500 epochs with a quicker coarse grid, then switch to a fine grid for 1500 epochs. The switch to the finer grid results in immediate uptick in loss, which quickly recovers. The convergence by epoch remains similar to non-multilevel approaches, but the overall time is reduced in the multilevel approach.}
			\label{fig:multi}
	\end{figure*}

	We test the multilevel strategy for the \dto{} and \otd{} approaches on the \textsc{Miniboone} data set (Figure~\ref{fig:multi}). 
	For both approaches, we train the first 500 epochs with a coarse grid and initial learning rate 1e-3 then the next 1500 epochs with a finer grid (using the ODE solver in Table~\ref{tab:flows}) while lowering the learning rate to 5e-4.
	For \dto{}, we start with a Runge-Kutta 4 scheme using $h{=}0.50$ then switch the step size to $h{=}0.25$. For \otd{}, we start with dopri5 with relative tolerance 1e-3 and absolute tolerance 1e-5, then switch to the default dopri5 (dividing each tolerance by $10^3$). For both approaches, when we switch grids at epoch 500, we witness a large uptick in the loss, which quickly decays. Overall, the performance of the multilevel schemes is comparable in both cases, and a similar reduction of the training time is observed in Figure~\ref{fig:multi}.

\section{Discussion} \label{sec:discussion}
	We compare the discretize-optimize (Disc-Opt) and optimize-discretize (Opt-Disc) approaches for time-series regression and continuous normalizing flows (CNFs) using neural ODEs.
	The \dto{} approach achieves similar or superior values of the validation loss at reduced computational costs, similar to the classification results in \citet{gholami2019anode}. 
	In our time-series regression example, \dto{} achieves an average 20x speedup over \otd{}. For the synthetic CNF problem of a two-dimensional Gaussian mixture model, \dto{} requires about half the number of function evaluations. 
	Similar results are observed on more realistic large-scale applications.
	Consider, e.g., the data set \textsc{Bsds300}. The \dto{} approach finished training in less than 14 hours, while the \otd{} trained for more than 8 days.
	In other data sets, runtime difference was less pronounced. However, the \dto{} method led to speedups on all except \textsc{Hepmass}, with an average speedup of 6x in total. Since the inverse error in this case is comparably small, we could perhaps use a larger time step or adaptive time stepping, directions for future work. We could also combine the numerical benefits of \dto{} with regularized neural ODEs, shown to speedup training 2.8x in the \otd{} approach~\citep{finlay2020train}. 
	
	One reason for these differences in training performance is the potential inaccuracies of the gradients in \otd{} approaches, as also addressed in~\citet{gholami2019anode}.
	We review this known issue by comparing backpropagation (used in \dto) to the adjoint method (used in \otd). 
	In the latter, gradients can become inaccurate when state and adjoint equations are not solved well enough~\citep{gholami2019anode}.
	In \dto, the gradients are accurate for any accuracy, allowing the user to guide solver accuracy by the accuracy of the data; e.g., the noise level in scientific applications. 
	We also show this property numerically.

	Our results are in line with~\citet{gholami2019anode}, who showed the improved convergence of \dto{} over the \otd{} approach on image classification tasks.
	Here, we show similar properties for time-series regression and CNFs using several numerical examples. 
	These applications differ from classification in that the inference in these applications requires the continuous model.
	For example, in CNFs, the inverse of the model is required for inference.
	The continuous model is invertible by design, which may motivate one to prefer \otd{} over \dto{}. 
	Remarkably, in our examples, \dto{} approaches can achieve competitive loss values and (upon re-discretization) low inversion errors.
	Notably, the \dto{} models in our examples were faster to train. 
	We also show that both approaches can be accelerated in similar ways using multilevel training. 
	
	In conclusion, we advocate for \dto{} approaches for training neural ODEs in time-series regression and continuous normalizing flows. 
	In our view, the inconvenience of choosing a sufficiently accurate time integration technique is compensated by the computational savings; since the \dto{} settings already exist in many codebases, implementation costs are minimal.
	Also, in our experiments, the models trained with \dto{} generalize with respect to re-discretization.
	
	\pagebreak 
	
\acks{Supported by the National Science Foundation award DMS 1751636, AFORS grant 20RT0237, Binational Science Foundation Grant 2018209, and a GPU donation by NVIDIA Corporation.}

\appendix

\section{ Derivation of Adjoint Equations} \label{app:adj_eqns}

	We provide derivations behind the continuous backpropagation~\eqref{eq:back_cont} and discrete  backpropagations~\eqref{eq:back_do} and~\eqref{eq:back_od}.

	\subsection{Continuous Adjoint}

	Consider the continuous ResNet optimization problem~\eqref{eq:Lreg} subject to~\eqref{eq:cont_ode}, 
	\begin{equation}
		\min_{\bfth} \, \int_0^T L \left( \bfy (t),\bfu(t) \right) \du t\quad
		\text{s.t.} \quad \partial_t \bfy(t) = \ell \big(\bfth(t),\bfy(t),t \big) , \quad \bfy(0) = \bfy_0
	\end{equation}
	where $\bfu$ is the ground truth and $\bfy$ depends on $\bfth$.
	
	From here, we calculate the Lagrangian $\calG$, with adjoint variable (and Lagrangian multiplier) $\bfz$
	\begin{equation} \label{eq:lagrange}
		\begin{aligned}
		\calG(\bfth,\bfy, \bfz) &= \int_0^T  L(\bfy(t),\bfu(t)) \du t\, + \, \int_0^T \bfz(t)^\top \, \big(  \ell (\bfth(t),\bfy(t),t) - \partial_t \bfy(t) \big) \du t.	\\	
		&= \int_0^T L(\bfy(t),\bfu(t)) \, + \,  \bfz(t)^\top \ell (\bfth(t),\bfy(t),t) - \bfz(t)^\top \partial_t\bfy(t) \, \, \du t . 
		\end{aligned}
	\end{equation}
	From optimization theory, we know that all the variations of $\calG$ have to vanish at an optimal point.
	Here, we derive the strong form of the optimality system.
	Before doing so, we simplify the third term using integration by parts and simplify using $\bfz(0)=0$ (due to the initial condition of the neural ODE):
	\begin{equation} \label{eq:parts}
		 \int_0^T  \bfz(t)^\top  \partial_t \bfy(t) \, \du t = \bfz(t)^\top \bfy(t) \biggr\rvert_0^T  - \int_0^T \partial_t \bfz(t)^\top \bfy(t)  \, \du t = \bfz(T)^\top \bfy(T) - \int_0^T \partial_t \bfz(t)^\top \bfy(t)  \, \du t .
	\end{equation}

	After substituting~\eqref{eq:parts} into~\eqref{eq:lagrange}, the Lagrangian now becomes
	\begin{align}
		\calG(\bfth,\bfy, \bfz)
		&= \int_0^T L(\bfy(t),\bfu(t)) \, + \,  \bfz(t)^\top \ell (\bfth(t),\bfy(t),t) + \partial_t \bfz(t)^\top  \bfy(t)  \, \du t \, - \, \bfz(T)^\top \bfy(T)  .
	\end{align}
	For some time $t\in[0,1)$, the variational derivative of $\calG$ with respect to $\bfy(t)$ is
	\begin{equation} \label{eq:adjoint}
		\begin{aligned}
		\partial_{\bfy(t)} \calG(\theta,\bfy,\bfz) &=  \nabla_{\bfy} L(\bfy(t),\bfu(t)) \, + \,   \nabla_{\bfy}  \ell (\bfth(t),\bfy(t),t) \, \bfz(t) + \, \partial_t \bfz(t) \\
		\end{aligned}
	\end{equation} 
	Setting this equal to zero gives the backward in time ODE
	\begin{equation}
	 -\partial \bfz(t) = L(\bfy(t),\bfu(t)) \, + \,  \nabla_{\bfy}  \ell (\bfth(t),\bfy(t),t) \bfz(t).
	\end{equation}
    Using that this equation holds for all $t \in [0,T)$, we see that the variational derivative of $\calG$ with respect to $\bfy(T)$ is
    \begin{equation}
        \nabla_{\bfy(T)} \calG(\theta,\bfy,\bfz) = \nabla_{\bfy} L(\bfy(T),\bfu(T)) - \bfz(T). 
    \end{equation}
    Setting this equal to zero gives the final time condition
    \begin{equation}
        \bfz(T) = \nabla_{\bfy} L(\bfy(T),\bfu(T))
    \end{equation}

Finally, we note that, via chain rule, the variation of $\calJ$ with respect to $\bfth$ is
\begin{equation}
    \nabla_{\bfth(t)} \calJ(\bfth) =   \nabla_{\bfth} \ell (\bfth(t),\bfy(t),t)\bfz(t).
\end{equation}

	\subsection{Discrete Adjoints}
	
	The backpropagations presented in~\eqref{eq:back_do} and~\eqref{eq:back_od} come from discretizing the continuous adjoint~\eqref{eq:back_cont}.
	
	We consider a deep discrete ResNet (forward Euler scheme) and calculate the \dto{} discretization for the backpropagation~\eqref{eq:back_do}. The forward mode~\eqref{eq:f_euler} follows
	\begin{equation*}
		\begin{aligned}
			\bfy_1 &= \bfy_0 + h \, \ell (\bfth_0,\bfy_0,t_0) \\
			\bfy_2 &= \bfy_1 + h \, \ell (\bfth_1,\bfy_1,t_1) \\
			\vdots \, \, \, & \quad  \qquad  \vdots \\
			\bfy_N &= \bfy_{N{-}1} + h \, \ell (\bfth_{N{-}1},\bfy_{N{-}1},t_{N{-}1}).
		\end{aligned}
	\end{equation*} 
	From outputs $\bfy_1,\dots,\bfy_N$, we calculate a discrete sum as the loss
	\begin{equation}
		J(\bfth) = \sum_{i=1}^N h \, L(\bfy_i,\bfu_i) . 
	\end{equation}
	
	We then backpropagate to calculate the gradients to update model parameters. Using auxiliary term $\bfz$ to accumulate the gradient of $J$ with respect to the states $\bfy$ as we step backwards in time,
	\begin{equation*}
		\begin{aligned}
			\bfz_N &= h\, \nabla_{\bfy_N} L(\bfy_N,\bfu_N) \\
			\bfz_{j} &= h\, \nabla_{\bfy_{j}} L(\bfy_{j},\bfu_{j})  
							+ h\, \nabla_{\bfy_{j}} L(\bfy_{j+1},\bfu_{j+1})\\
			&= h\, \nabla_{\bfy_{j}} L(\bfy_{j},\bfu_{j})  
							+ \nabla_{\bfy_{j}} \bfy_{j+1}  \, \, \, h\, \nabla_{\bfy_{j+1}} L(\bfy_{j+1},\bfu_{j+1}) \\
			&= h\, \nabla_{\bfy_{j}} L(\bfy_{j},\bfu_{j})  
							+ (I + h \nabla_{\bfy_{j}}\ell(\bfth_{j},\bfy_{j},t_{j})) \, \bfz_{j+1} \\
			&= \bfz_{j+1} + h \big( \nabla_{\bfy}\ell(\bfth_j,\bfy_j, t_j) \bfz_{j+1} + \nabla_{\bfy}\regL(\bfy_j, \bfu_j) \big)
		\end{aligned}
	\end{equation*}

	Using chain rule, we can now calculate the gradient with respect to the parameters
	\begin{equation}
		\begin{aligned}
			\nabla_{\bfth_j} J(\bfth) &= \nabla_{\bfth} \, \bfy_j \, \, \nabla_{\bfy_j} J(\bfth) \\
			&= h \nabla_{\bfth} \, \ell(\bfth_j,\bfy_j,t_j) \, \, \bfz_j .
		\end{aligned}
	\end{equation}

	The \otd{} discretization~\eqref{eq:back_od} follows the backward Euler scheme. It calculates the gradient at time $t_{j{+}1}$ instead of at $t_{j}$ as \dto{} does.

\vskip 0.2in
\bibliography{main.bib}

\end{document}